\documentclass{article}

\usepackage[preprint]{neurips_2024}

\usepackage[utf8]{inputenc} % allow utf-8 input
\usepackage[T1]{fontenc}    % use 8-bit T1 fonts
\usepackage{hyperref}       % hyperlinks
\usepackage{url}            % simple URL typesetting
\usepackage{booktabs}       % professional-quality tables
\usepackage{amsfonts,amsmath,amssymb,amsthm} % blackboard math symbols
\usepackage{nicefrac}       % compact symbols for 1/2, etc.
\usepackage{microtype}      % microtypography
\usepackage{graphicx}
\usepackage{float}
\usepackage{newfloat}
\usepackage{wrapfig,framed}
\usepackage{tasks}
\usepackage{array}
\usepackage{bbm}
\usepackage{soul}

\usepackage{times}
\usepackage{soul}
\usepackage[utf8]{inputenc}
\usepackage{caption}
\usepackage{algorithm}
\usepackage{algorithmic}
\usepackage{footnote}
\newtheorem{lemma}{Lemma}
\usepackage{booktabs}
\usepackage{listings}
\usepackage{paralist}
\usepackage{multirow}
\usepackage[table]{xcolor}
\usepackage{subcaption}
\usepackage{comment}

\floatstyle{ruled}
\newfloat{listing}{tb}{lst}{}
\floatname{listing}{Listing}
\newcommand*{\Scale}[2][4]{\scalebox{#1}{$#2$}}%
\newcolumntype{C}[1]{>{\centering\arraybackslash}m{#1}}

\title{Designing User-Centric Behavioral Interventions to Prevent Dysglycemia with Novel Counterfactual Explanations}
\author{%
  Asiful Arefeen, Hassan Ghasemzadeh\\
  Arizona State University\\
}

\begin{document}

\maketitle

\begin{abstract}
Monitoring unexpected health events and taking actionable measures to avert them beforehand is central to maintaining health and preventing disease. Therefore, a tool capable of predicting adverse health events and offering users actionable feedback about how to make changes in their diet, exercise, and medication to prevent abnormal health events could have significant societal impacts. Counterfactual explanations can provide insights into why a model made a particular prediction by generating hypothetical instances that are similar to the original input but lead to a different prediction outcome. Therefore, counterfactuals can be viewed as a means to design AI-driven health interventions to not only predict but also prevent adverse health outcomes such as blood glucose spikes, diabetes, and heart disease. In this paper, we design \textit{\textbf{ExAct}}, a novel model-agnostic framework for generating counterfactual explanations for chronic disease prevention and management. Leveraging insights from adversarial learning, ExAct characterizes the decision boundary for high-dimensional data and performs a grid search to generate actionable interventions. ExAct is unique in integrating prior knowledge about user preferences of feasible explanations into the process of counterfactual generation. ExAct is evaluated extensively using four real-world datasets and external simulators. With $82.8\%$ average validity in the simulation-aided validation, ExAct surpasses the state-of-the-art techniques for generating counterfactual explanations by at least $10\%$. Besides, counterfactuals from ExAct exhibit at least $6.6\%$ improved proximity compared to previous research.
\end{abstract}

\section{Introduction}
Unhealthy lifestyle contributes to chronic complications like diabetes and cardiovascular disease and increases risks of morbidity and mortality \cite{Lakka2002TheMS, Nathan2005IntensiveDT}. However, preventing these abnormal events through lifestyle changes is challenging due to ingrained habits, societal influences, and the difficulty in sustaining long-term behavior modifications. Wearables technologies (e.g., CGMs, smartwatches, insulin pumps) can continuously monitor dietary intake, exercise, medication behavior and AI algorithms can predict abnormal events ahead of time leveraging sensor data, thus providing opportunities to prevent such events. However, it is currently unknown how exactly predictions made by machine learning algorithms can be used to modify human behavior to prevent adverse outcomes. To address this knowledge gap, we design a novel approach to generate counterfactual (CF) explanations to reason about the model predictions and determine minimum behavioral modifications that a person can make to avert abnormal health events.

CF explanations research is closely related to explainable AI (XAI). The adoption of XAI is gaining momentum across different domains, extending beyond conventional feature-importance methods and embracing techniques including causal, visual and CF explanations for instilling trust and fairness in black-box models \cite{
ChapmanRounds2021FIMAPFI}. While technologies like LIME \cite{Ribeiro2016WhySI}, TIME \cite{Sood2021FeatureIE}, and SHAP \cite{Lundberg2017AUA} focus on creating a hierarchy of features based on their contributions to the model's prediction, explanatory tools such as Grad-CAM \cite{Selvaraju2016GradCAMVE} and SmoothGrad \cite{Smilkov2017SmoothGradRN} provide visual interpretations highlighting the specific segments within continuous data that contribute the most towards a specific class. CF explanations is a more targeted branch of XAI that emphasizes describing the smallest change to the feature values that changes the prediction outcome to a predefined output. CF are not necessarily actual instances from the training data, but can be adversarial samples made with a combination of feature values. Prior research \cite{Wachter2017CounterfactualEW, mothilal2020dice, Dandl2020MultiObjectiveCE, Navas-Palencia2021Counterfactual, Karimi2019ModelAgnosticCE} formulated CF explanations generation as a multi-objective optimization problem (MOOP) and solved it leveraging different optimizers, including ADAM \cite{Wachter2017CounterfactualEW}, SGD \cite{Dandl2020MultiObjectiveCE}, NSGA-II \cite{Dandl2020MultiObjectiveCE,Navas-Palencia2021Counterfactual}, and SAT solver \cite{Karimi2019ModelAgnosticCE}, respectively. In contrast, DACE \cite{2020DACEDC} generated CF explanations based on mixed integer linear optimization with a custom loss function. Several researchers leveraged iterative steps to modify the actionable features greedily until the predicted class is changed \cite{Gomez2020ViCEVC, Schut2021GeneratingIC}. NICE \cite{Brughmans2021NICEAA} generates plausible CFs similar to the original instance by searching for nearby instances in the data manifold that exhibit the desired outcome. However, prior research falls short in integrating the user's ranking of plausible features into the optimization process for generating CF explanations. DiCE \cite{mothilal2020dice} incorporates user-defined constraints on certain features by formulating the creation of CF explanations as a mixed-integer optimization problem. However, DiCE first generates counterfactuals without considering the constraints and then enforces some constraints through post-processing to ensure certain preferences are met.

CF explanations serve as a way to design clinical interventions. While designing behavior change interventions, focusing on small changes is crucial to ensure sustainability of the behavior change \cite{Arefeen2022ComputationalFF} - a key concept from both Social Cognitive theory \cite{Bandura1985SocialFO} and CF explanations. While minimum change is necessary in behavioral intervention, prioritizing user preferences in the selection of the modifiable features is also a key attribute for successful adoption of the AI system. However, such effective, sustainable and user-focused interventions have not been studied in CF explanation research. ExAct bridges these gaps by introducing a model-agnostic explainable AI intervention system for generating personalized CF explanations for chronic disease prevention and management. Contributions made by ExAct can be summarized as follows-
\begin{itemize}
    \item Using the predictive model, ExAct devises a new technique to generate CF explanations by approximating the decision boundary in high-dimensional data followed by a grid search.
    \item It outlines a minimum change based behavioral modification scheme that is grounded in well-known theories in neuro-psychology.
    \item Another scheme is also designed which reflects user preference to keep certain features unchanged of a test point in a high-dimensional space. ExAct models user preferences as soft constraints using a preference ranking approach.
\end{itemize}

\section{Problem statement}

Let $\mathcal{D}$ = \{($X_1$,$y_1$), ($X_2$, $y_2$), $\dots$ , ($X_n$, $y_n$)\} be a dataset of $n$ instances that captures longitudinal health data and corresponding health outcome such as blood glucose, occurrence of diabetes, risk of heart disease. Each instance $X_i$ = [${x_i}^{1}$, ${x_i}^{2}$, $\dots$, ${x_i}^{d}$] consists of $d$ features including actionable behavioral parameters (e.g., exercise, diet, medication) and non-actionable parameters (e.g., age, gender). Considering $c$ possible classes for health outcome $Y$, where $y_i \in [1,c]$, a probabilistic AI model or classifier $f$ can be trained to map the $d$-dimensional input features to the $c$ classes:
\vspace{-0.5mm}
\begin{equation}
f : \mathbb{R}^{d} \rightarrow \mathbb{R}^{c}
\end{equation}
\vspace{-0.5mm}
As part of the probabilistic model, \textit{softmax} activation is applied at the last layer, generating a probability score for each class. Therefore, for a sample $X \in \mathbb{R}^{d}$, the assigned class is denoted as $\hat{y} = \underset{y}{\mbox{\textit{argmax}}}f(X)$, and the corresponding class score or probability for $\hat{y} = a$ is $f_a$($X$) = $P$($y=a \mid X$) = \textit{max}$\big(f(X)\big)$. The predicted class can be categorized as normal (e.g., normoglycemia, healthy etc.) and abnormal (e.g. hyperglycemia,  hypoglycemia, sleep disorder, diabetic etc.) in the problem under study in this paper.

%As part of the probabilistic model or classifier, softmax activation is applied at the last layer, generating a probability score for each class. Therefore, for any sample $X \in \mathbb{R}^{d}$, the assigned class is denoted as $\hat{y} = \underset{a}{\mbox{argmax}}f(X)$, and the corresponding class score or probability is $f_a(X) = P(y=a|x)=max\big(f(X)\big)$. The predicted class can be categorized as normal, abnormal, or even atypical, such as relaxed, stressed, or moderate in stress management.

If a test sample $X_T$ is predicted to indicate an abnormal health event (e.g., ${\mbox{argmax}}f(X_T)$ = $abnormal$), an important question is how to provide an intervention plan to assist the patient with making appropriate behavior changes to prevent the impending abnormal event while considering user's preferences at the same time. We assume that the user's preferences for behavior changes are represented in vector $R(X_T)$ = \{$r_1$, $\dots$, $r_d$\} where each $r_i \in$ \{$1$, $\dots$, $d$\} indicates the ranking of the $i$-th feature for modification during intervention. For example, if $r_4 = 1$, it means that the user ranks the fourth feature as the most favored behavioral factor for modification. Therefore, our goal in ExAct is to create CF explanations $(X_T^*)$ that satisfy the following requirements: (i) \textit{interventional}: $X_T^*$ must change the class of the initial prediction from an abnormal class to the normal class; (ii) \textit{minimal}: $X_T^*$ must be minimally distant from the factual sample $X_T$; (iii) \textit{realistic}: $X_T^*$ must look realistic, i.e., the features of the CFs must fall within the distribution of the dataset $\mathcal{D}$; (iv) \textit{partial}: $(X_T^*)$ must favor user preferences expressed in feature rank $R$. The CF generation process can be formulated as shown in \eqref{eq:cfxf} where the interventional, minimal, realistic, and partial requirements are formalized, respectively, in the first to the fourth terms in the minimization problem.
\vspace{-1.3mm}
\begin{equation}
\begin{aligned}
\label{eq:cfxf}
\min_{X_T^*}\Big[CE\big(f(X_T^*),\overrightarrow{n}\big)+ d\big(X_T^*,X_T\big) + d\big(X_T^*,X|_{\text{target class}}\big) + R(X_T)\odot|X_T^*-X_T| \Big]
\end{aligned}
\end{equation}
\vspace{2mm}
In \eqref{eq:cfxf}, the term $CE(\cdot)$ represents the crossentropy loss calculated on model's prediction on the CF and \textit{normal} class. Optimizing for this term addresses the \textit{interventional} requirement while optimizing for $d\big(X_T^*,X_T\big)$ ensures that $X_T^*$ remains in close proximity of the factual instance. Minimizing $d\big(X_T^*,X\big)$ results in keeping the feature values of $X_T^*$ within range and ensures \textit{realism}. The last term in \ref{eq:cfxf} implies that changes to features which the user prefers to modify (higher importance) are weighted more heavily in the optimization process. It encourages the optimization process to prioritize changes in those features by weighting the distance with the user preference vector and therefore satisfies the \textit{partial} requirement. We will dissect the optimization into the sub-problems below.

\textbf{Problem 1 (Accessing the decision boundary)} One approach to produce CFs is by accessing the decision boundary of the trained model. Samples close to the decision boundary are minimally distant from $X_T$, hold the opposite class, have feature values within range and thus satisfies most of the requirements in Equation~\ref{eq:cfxf}. Therefore, accessing the decision boundary could be a reasonable proxy to optimizing the objective function shown in Equation~\ref{eq:cfxf}. The classifier $f$ divides the space $\mathbb{R}^d$ into $c$ decision regions, represented as $\mathfrak{r}_1, \mathfrak{r}_2, ..., \mathfrak{r}_c$. The decision boundaries are thin regions where the classifier is most uncertain about the label to be assigned to a sample. Hence, mathematically, the decision boundary between classes $a$ and $b$ can be defined as $\mathfrak{r}_{a,b} = {v \in \mathbb{R}^d : P(y=a|x) = P(y=b|x)}$. To realize this approach to generating CF explanations, first, we must construct the decision boundary of the classifier trained on a high-dimensional feature set.

\textbf{Problem 2 (Designing intervention)} Even if the high-dimensional decision boundary of the classifier is made accessible, how do we design an intervention that ensures minimum change based modification and/or prioritizes user preferences? This means that, if a test sample $X_T = {{x_T}^{1}, {x_T}^{2}, ..., {x_T}^{d}}$ is predicted to be \textit{abnormal}, i.e. ${\mbox{argmax}}f(X_T) = abnormal$, one needs to find a path for transitioning from the \textit{abnormal} factual instance to the \textit{normal} region by leveraging the decision boundary of the classifier. To this end, the intervention design algorithm involves performing a series of operations on the test sample such that the predicted class is altered, i.e. ${\mbox{argmax}}f(X_T^*) = normal $ and the user preference requirement is entertained.

\begin{figure*}[!h]
    \centering
    \subfloat[\label{a}]{{\includegraphics[scale=0.43]{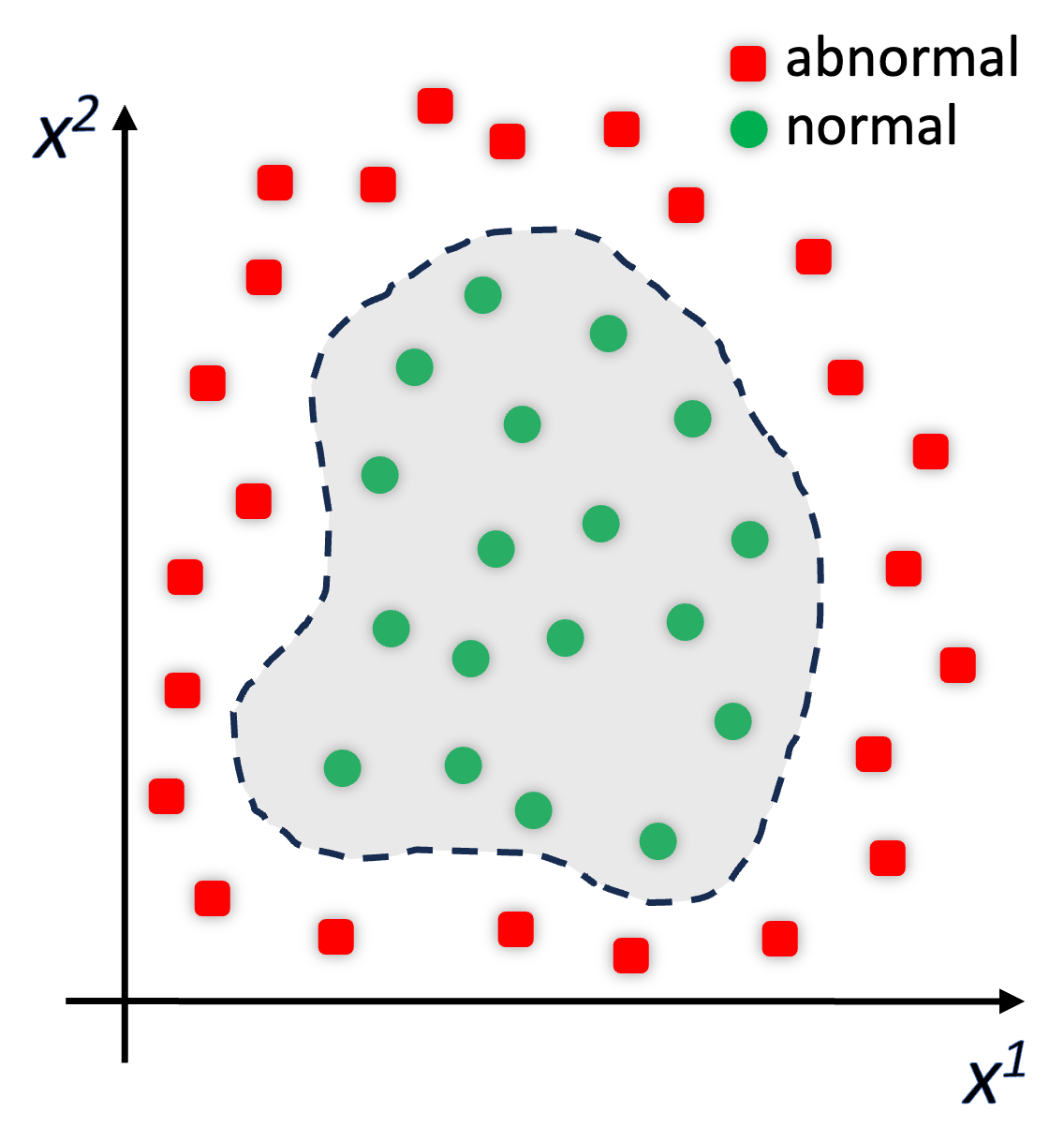}}}%
    \quad
    \subfloat[\label{b}]{{\includegraphics[scale=0.43]{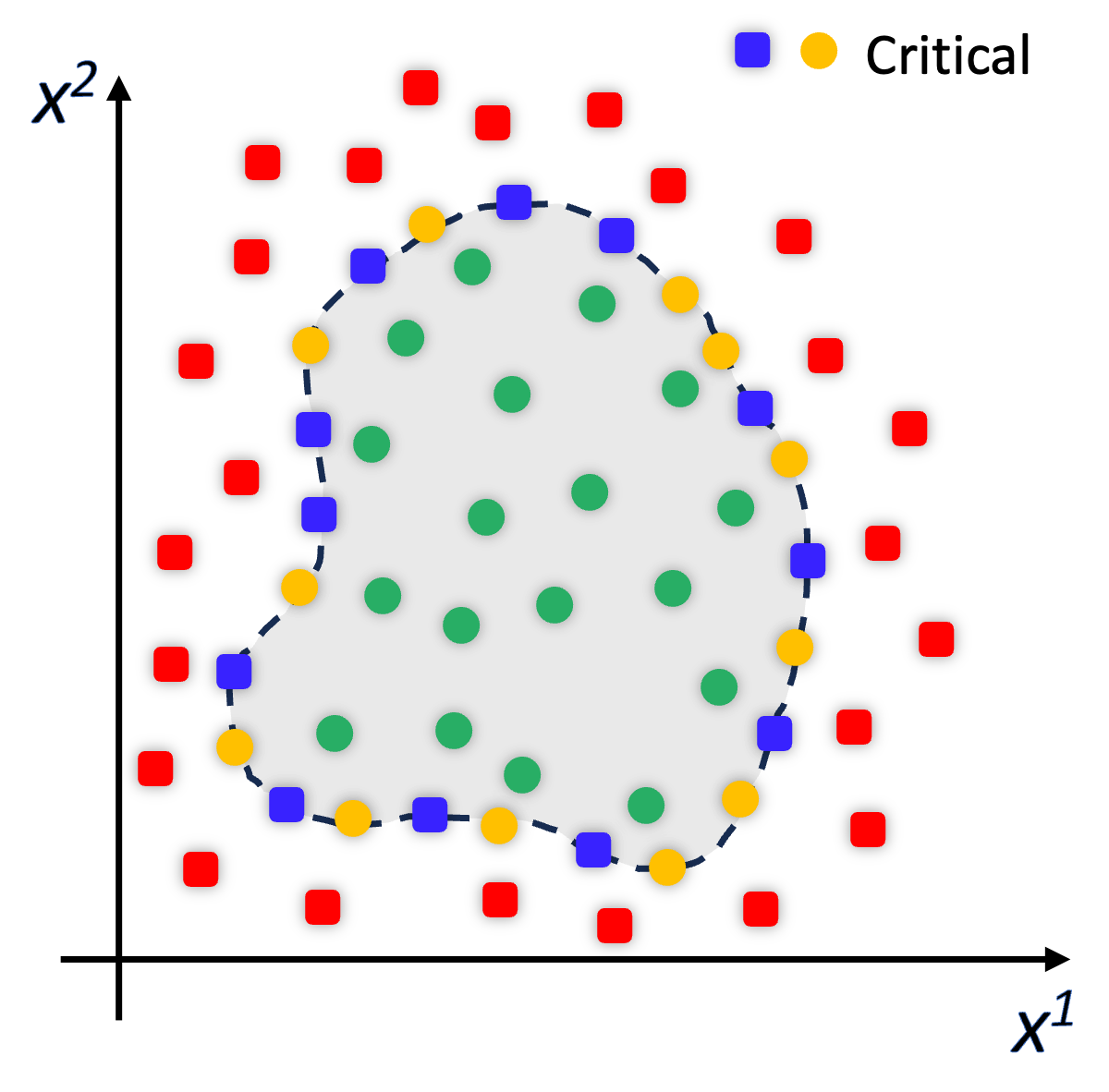}}}%
    \quad
    \subfloat[\label{c}]{{\includegraphics[scale=0.43]{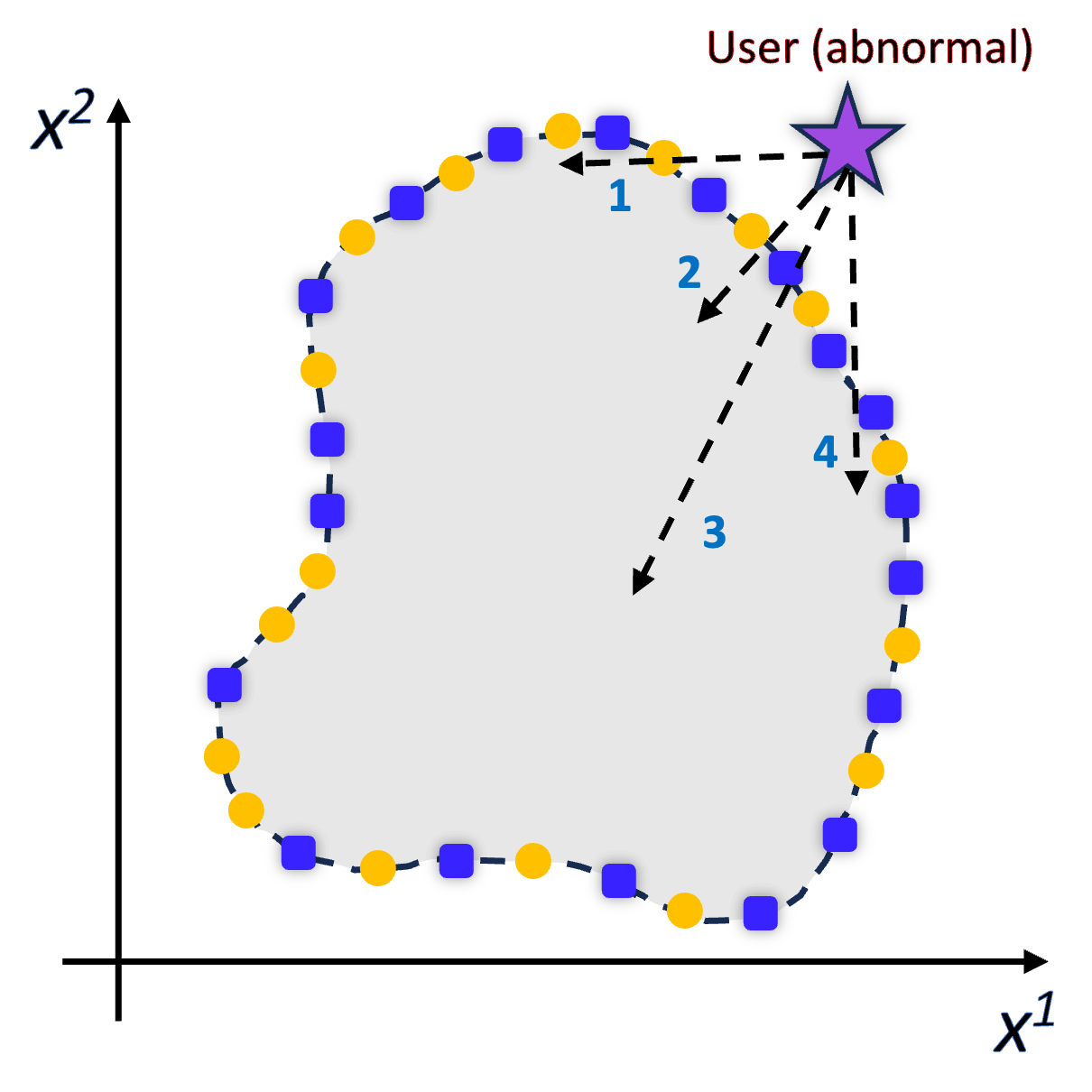}}}%
    \caption{Proposed approach to solve the optimization problem. (a) A classifier (black dashed line) is trained to identify \textit{normal} (green) vs. \textit{abnormal} (red) classes. (b) Many adversarial critical samples (borderline instances in blue and yellow) are generated along the decision boundary to approximate the decision hyperplane. (c) The real samples can be removed once substantial number of critical samples are in place. For a test sample outside of the \textit{normal} region, predicted class can be toggled following an infinite number of trajectories. However, \textit{ExAct} follows the path that requires \textbf{minimal changes} or emphasizes \textbf{users preferences} and restricts some features while flipping the class.}
    \label{o_view}%
\end{figure*}

\section{Problem solution}
Figure~\ref{o_view} shows an overview of the proposed solution. Given the classifier defines specific regions for the classes using decision hyper-plane, accessing it is critical in producing CFs. Once access to the decision hyperplane is gained, as in Figure~\ref{c}, an effective intervention plan can be outlined for users who fall into the abnormal region (Figure~\ref{c}) for an improved health outcome.

\textbf{\textit{Approximating decision boundary}} Linear models are limited to linear decision boundaries and they are easily accessible. Being highly non-linear and intricate, decision boundaries of neural networks cannot be defined easily using simple equations \cite{GoodBengCour16,Bishop2006PatternRA,Hastie2005TheEO}. Instead, the decision boundaries are implicitly defined by the network's weights and activation functions, making their representation more challenging. Therefore, like Figure~\ref{b}, one possible approach is to use adversarial borderline samples as proxies to approximate the decision boundary.

To uncover the critical borderline instances, an autoencoder-driven technique \cite{Karimi2019CharacterizingTD} is devised. While a vanilla autoencoder emphasizes on minimizing the disparity between its input and output, its capabilities need to be augmented to generate critical adversarial samples along the decision boundary. When fed with a \textit{normal} sample $X_n$, the autoencoder $AE_1$ would reconstruct an output $AE_1(X_n)$ that closely resembles the input. At the same time, the anticipation is that this reconstructed sample, despite its similarity to the \textit{normal} input, will be classified as \textit{abnormal} by the predictive model $f$. To achieve this delicate balance, the loss function takes on a refined form-
\vspace{-1mm}
\begin{equation}{\label{AE1}}
\Scale[0.82]{\mathcal{L}_1 = \sum_{\forall X_n}\min\bigg[||X_{n}-AE_1(X_{n})||^2+\alpha \times CE \Big(f\big(AE_1(X_n)\big), \overrightarrow{a}\Big)\bigg]}
\end{equation}
where, $CE(\cdot)$ applies crossentropy on the model's prediction and $\overrightarrow{a}$. $\overrightarrow{a}$ is a one-hot encoded vector for \textit{abnormal} class. The hyperparameter $\alpha$ plays a crucial role in balancing between the reconstruction loss and the generation of adversarial examples. A similar loss function can be defined for the opposite process i.e. producing critical samples by feeding \textit{abnormal} samples to another autoencoder-
\begin{equation}{\label{AE2}}
\Scale[0.82]{\mathcal{L}_2 = \sum_{\forall X_a}\min\bigg[||X_{a}-AE_2(X_{a})||^2+\alpha \times CE \Big(f\big(AE_2(X_a)\big), \overrightarrow{n}\Big)\bigg]}
\end{equation}
\begin{figure}[!tbh]
\centering
\includegraphics[width=0.32\linewidth]{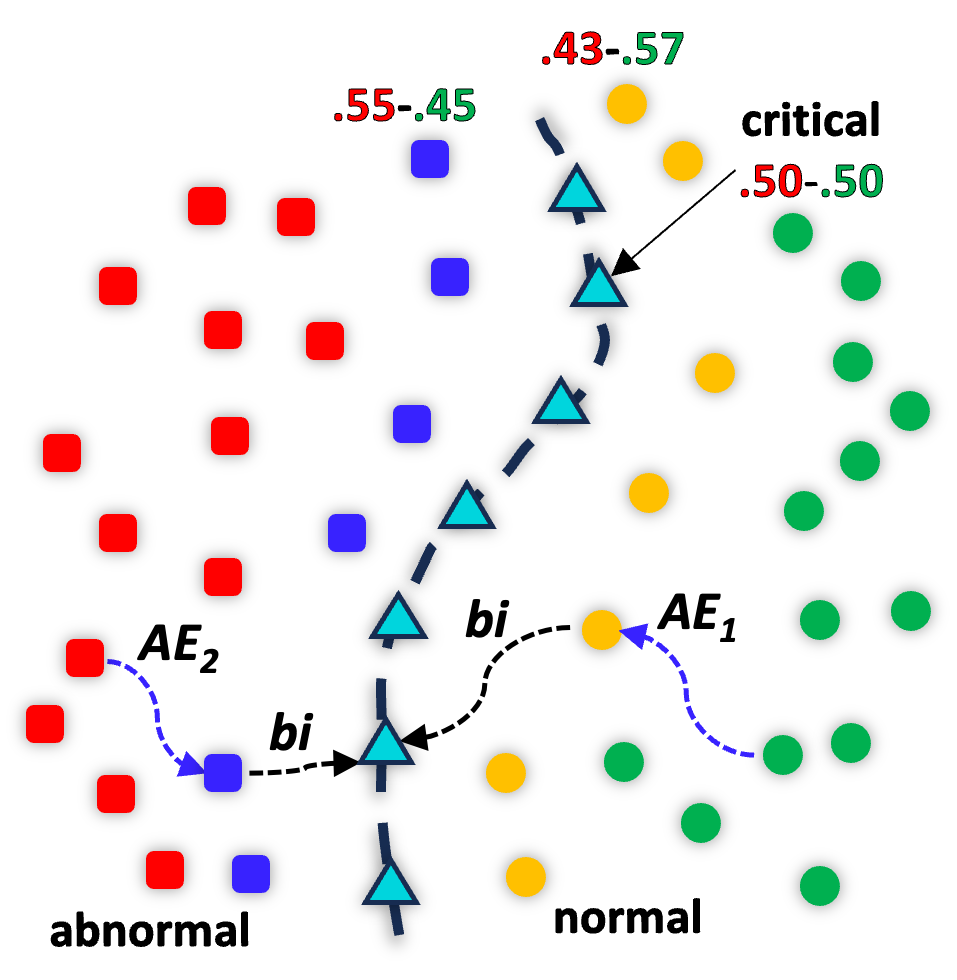}
\caption{The autoencoders produce borderline instances (blue and yellow), while the addition of a bisection algorithm adds finesse to them (teal triangles). The numbers are class probabilities.} 
\label{boundary}
\end{figure}
Although the aforementioned autoencoders attempt to produce borderline instances, the resulting adversarial samples may still fall considerably distant from the actual decision boundary. As pictured in Figure~\ref{boundary}, the blue and yellow samples, coming off the autoencoders, hold roughly 55-45\% probability. Therefore, to address this limitation, a \textit{bisection} algorithm comes into play to generate even more critical instances positioned precisely on the decision boundary with a balanced prediction probability of 50-50 (teal triangles in Figure~\ref{boundary}). The bisection method is illustrated in Algorithm~\ref{bisection_alg}.

\begin{algorithm}[!ht]
\footnotesize
\caption{Bisection method to identify the critical borderline instances}
\label{bisection_alg}
\textbf{Input}: close-proximity output pair\footnotemark from autoencoders $\hat{X}_n$ and $\hat{X}_a$,
predictive classifier $f$, threshold $\beta$ \\
\textbf{Output}: borderline instance $X_m$ \\
\textbf{Initialization}: $X_l = \hat{X}_n$, $X_r = \hat{X}_a$
\begin{algorithmic}[1] %[1] enables line numbers
\WHILE{condition}
\STATE $X_m = \frac{1}{2}(X_l+X_r)$
\IF {$f_n(X_m)>f_a(X_m)$}
\STATE $X_l = X_m$
\ELSIF{$f_n(X_m)<f_a(X_m)$}
\STATE $X_r = X_m$
\ENDIF
\IF {$|f_n(X_m)-f_a(X_m)|\leq\beta$}
\RETURN $X_m$
\ENDIF
\ENDWHILE
\end{algorithmic}
\end{algorithm}
\vspace{-1mm}
\textit{\textbf{Developing an intervention plan}} Figure~\ref{c} demonstrate the Rashomon effect associated with CF explanations i.e. there could be an infinite number of possible pathways for any \textit{abnormal} sample to penetrate the decision boundary and generate a CF \cite{Dandl2024CountARFactualsG}. However, it is important to note that not all of these are feasible or practical. For instance, expecting an elderly patient to suddenly increase their physical activity significantly may not be realistic. As stated previously, it is desirable to accommodate user preferences or adhere to doctor's recommendations by keeping certain aspects (e.g. protein or medication intake) unchanged. Therefore, the objective is to identify the most effective intervention plan that ensures \textbf{\textit{minimal changes}} or emphasizes \textbf{\textit{user preferences}}, ensuring a successful transition towards the \textit{normal} region.

\textbf{\textit{Minimal intervention:}} For the test sample $X_T \in \mathbb{R}^{d}$ that is classified as \textit{abnormal}, the aim is to find the smallest adversarial perturbation $\delta_{min,p}$, that can be added to the original test point $X_T$, such that the perturbed point $X_T +\delta$ remains within a specified set of constraints $C$ and the classifier decision changes to $normal$. Therefore, the minimal adversarial perturbation for $X_T$ with respect to the $l_p$-norm can be defined mathematically-
\begin{align*}
    \delta_{min,p} = \min_{\delta \in \mathbb{R}^d}||\delta||_p 
    \text{ subject to  }f_n(X_T +\delta)>\max_{a\neq n, \forall a} f_a(X_T +\delta),\text{ } X_T +\delta\in C
\end{align*}
\vspace{-0.5mm}
Since the decision boundary is approximated using critical borderline instances, the search for minimum perturbation narrows down to a search of minimum distance between the test point and a set of confusing or critical points. Therefore, the search can be illustrated further with the "Distance to Set Lemma" that relates to the expression- 
\begin{equation}\label{infimum}
    d(X_T, S) = \inf\{d(X_T, y) : y \in S\}
\end{equation}
Finding the minimum distance from test point $X_T$ to a set $S$ requires comparing the distance between $X_T$ and each point in $S$ and selecting the minimum. If the dimensionality of the space is fixed, the complexity can be considered as $\mathcal{O}(|S|)$, with $|S|$ being the cardinality of the set. 
\footnotetext{categorical features do not enter into the bisection algorithm.}
To achieve \textbf{\textit{minimal intervention}}, an approach to address Eqn.~\ref{infimum} involves populating the decision boundary with an abundance of adversarial samples and finding the one closest to $X_T$ by leveraging normalized nearest neighbor algorithm-
\begin{equation}
X_T^* = \underset{T}{\mbox{argmin}} \sum_{i=1}^{|S|}\sqrt{\sum_{j=1}^{d}(\Tilde{{x_T}^j}-\Tilde{s^j})^2}
\end{equation}
where, $\Tilde{{x_T}^j}$ and $\Tilde{{s}^j}$ are features normalized by $L_2$ norm of corresponding sample. Trajectory 2 in Figure~\ref{c} is the approximated representation of minimal intervention. 
\begin{lemma}\label{lemma1}
For a point $x$ and a non-empty set $S$ in a metric space, the distance from $x$ to $S$, denoted as $d(x, S)$, is equal to the infimum of the distances between $x$ and all points in $S$ i.e. $d(x, S) = \inf\{d(x, y) : y \in S\}$
\end{lemma}
\vspace{-1mm}
\begin{proof}
\textbf{Upper Bound:}
First thing to show is $d(x, S) \leq \inf\{d(x, y) : y \in S\}$. Let $\varepsilon > 0$ be arbitrary. By the definition of \textit{infimum}, there exists a point $y_\varepsilon$ in $S$ such that $d(x, y_\varepsilon) < \inf\{d(x, y) : y \in S\} +\varepsilon$. This implies that $d(x, y_\varepsilon) -\varepsilon < \inf\{d(x, y) : y \in S\}$.
Since $y_\varepsilon \in S$, it follows that $\inf\{d(x, y) : y \in S\} \leq d(x, y_\varepsilon)$. Combining the inequalities, we have $d(x, S) \leq d(x, y_\varepsilon) < \inf\{d(x, y) : y \in S\} +\varepsilon$. Since $\varepsilon$ was arbitrary, it concludes that $d(x, S) \leq inf\{d(x, y) : y \in S\}$.

\noindent\textbf{Lower Bound:}
Next thing to show is $d(x, S) \geq \inf\{d(x, y) : y \in S\}$. Let $\varepsilon > 0$ be arbitrary. By the definition of \textit{infimum}, there exists a point $y_\varepsilon$ in $S$ such that $d(x, y_\varepsilon) < \inf\{d(x, y) : y \in S\} +\varepsilon$.
Since $y_\varepsilon \in S$, we have that $d(x, y_\varepsilon) \geq \inf\{d(x, y) : y \in S\}$. Rearranging the inequality, we obtain $d(x, y_\varepsilon) -\varepsilon \geq \inf\{d(x, y) : y \in S\}$. Since $\varepsilon$ was arbitrary, it follows that $d(x, S) \geq \inf\{d(x, y) : y \in S\}$.

Combining the upper and lower bounds, it can be concluded that $d(x, S) = \inf\{d(x, y) : y \in S\}$.
\end{proof}
\begin{figure*}[!t]
    \centering
    \subfloat[\label{ia}]{{\includegraphics[scale=0.37,trim=2mm 2mm 2.5mm 2mm, clip]{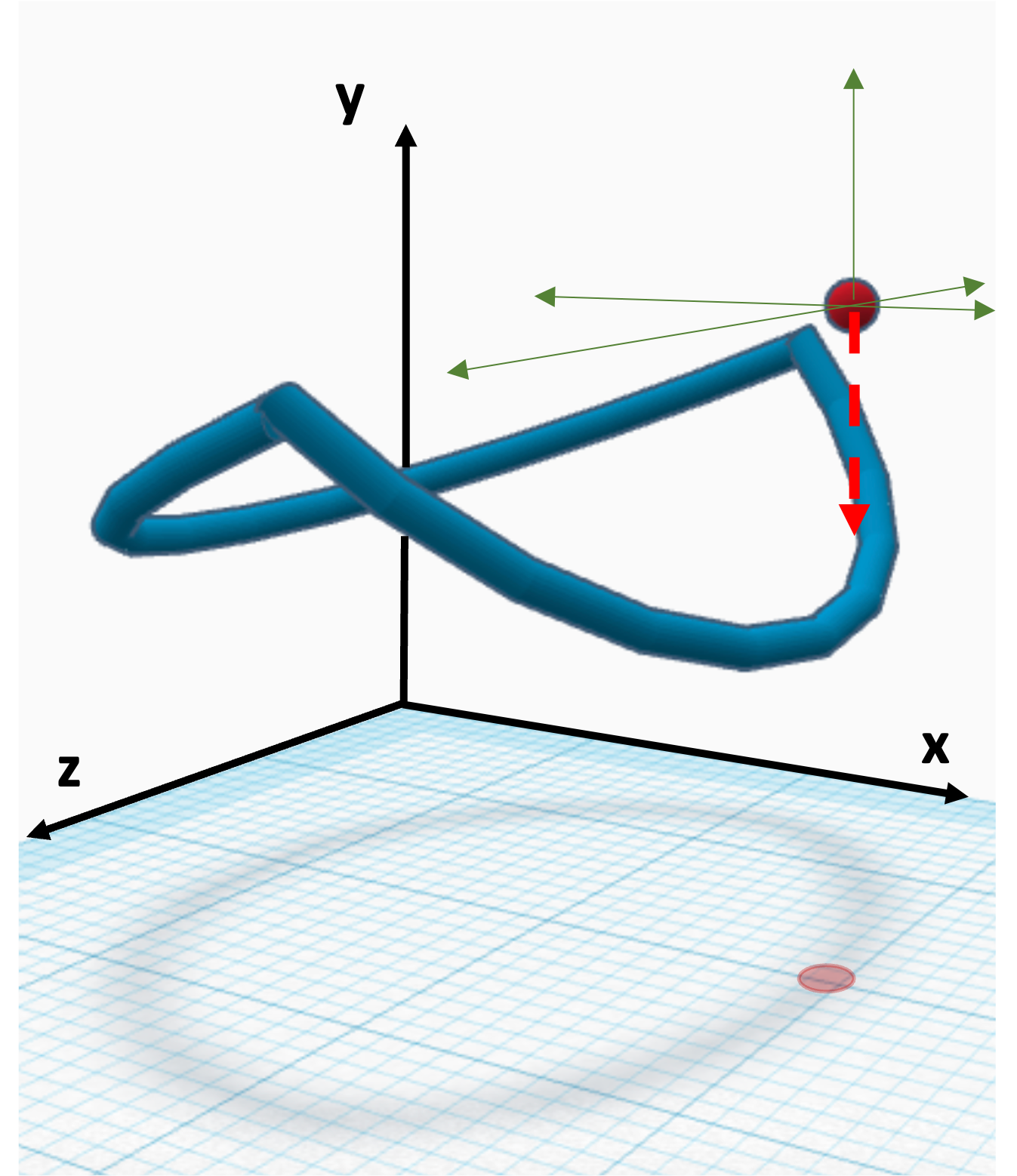} }}%
    \quad
    \subfloat[\label{ib}]{{\includegraphics[scale=0.39,trim=2mm 2mm 2mm 2mm, clip]{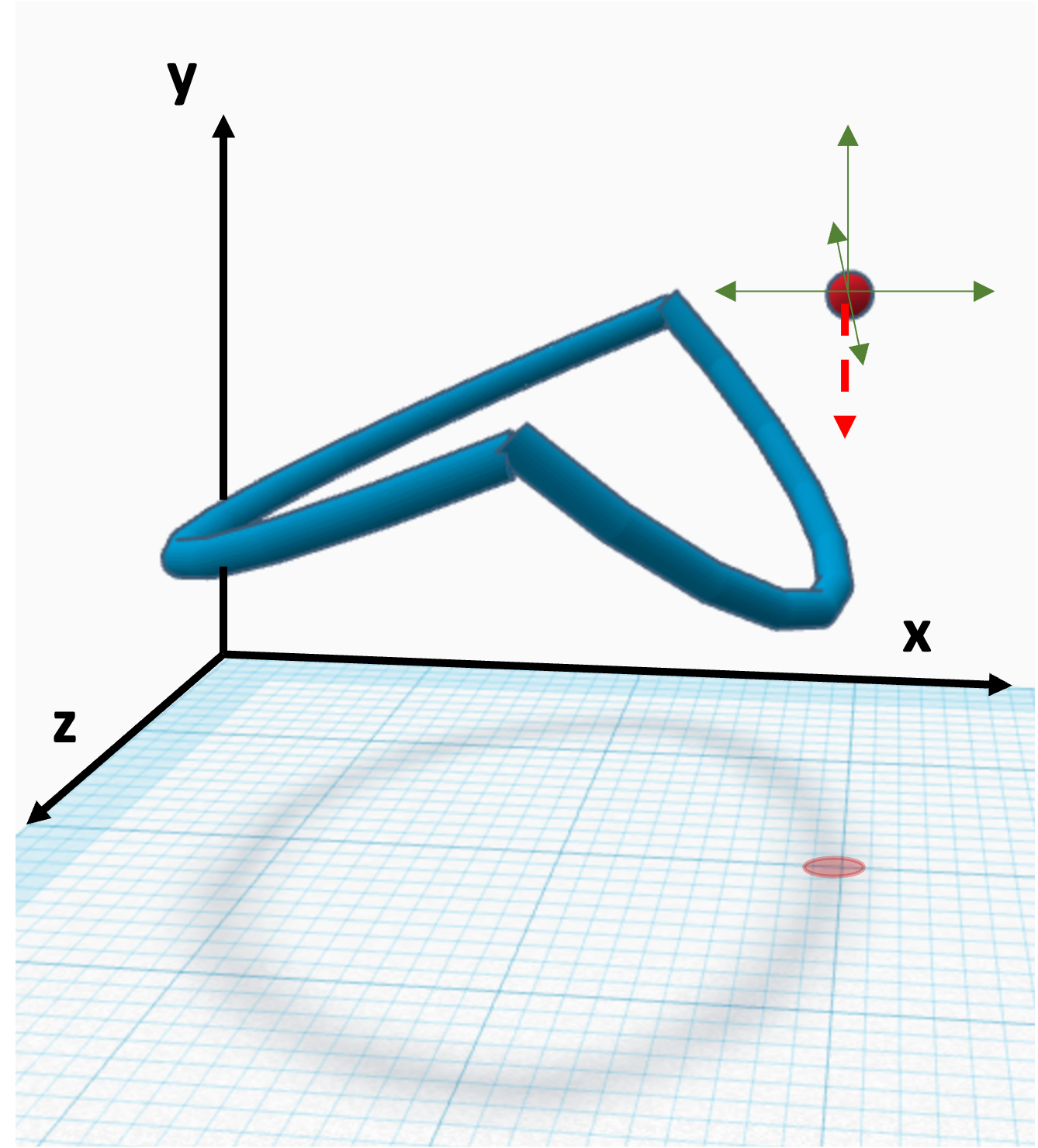} }}%
    \quad
    \subfloat[\label{ic}]{{\includegraphics[scale=0.348,trim=2mm 2mm 2mm 2mm, clip]{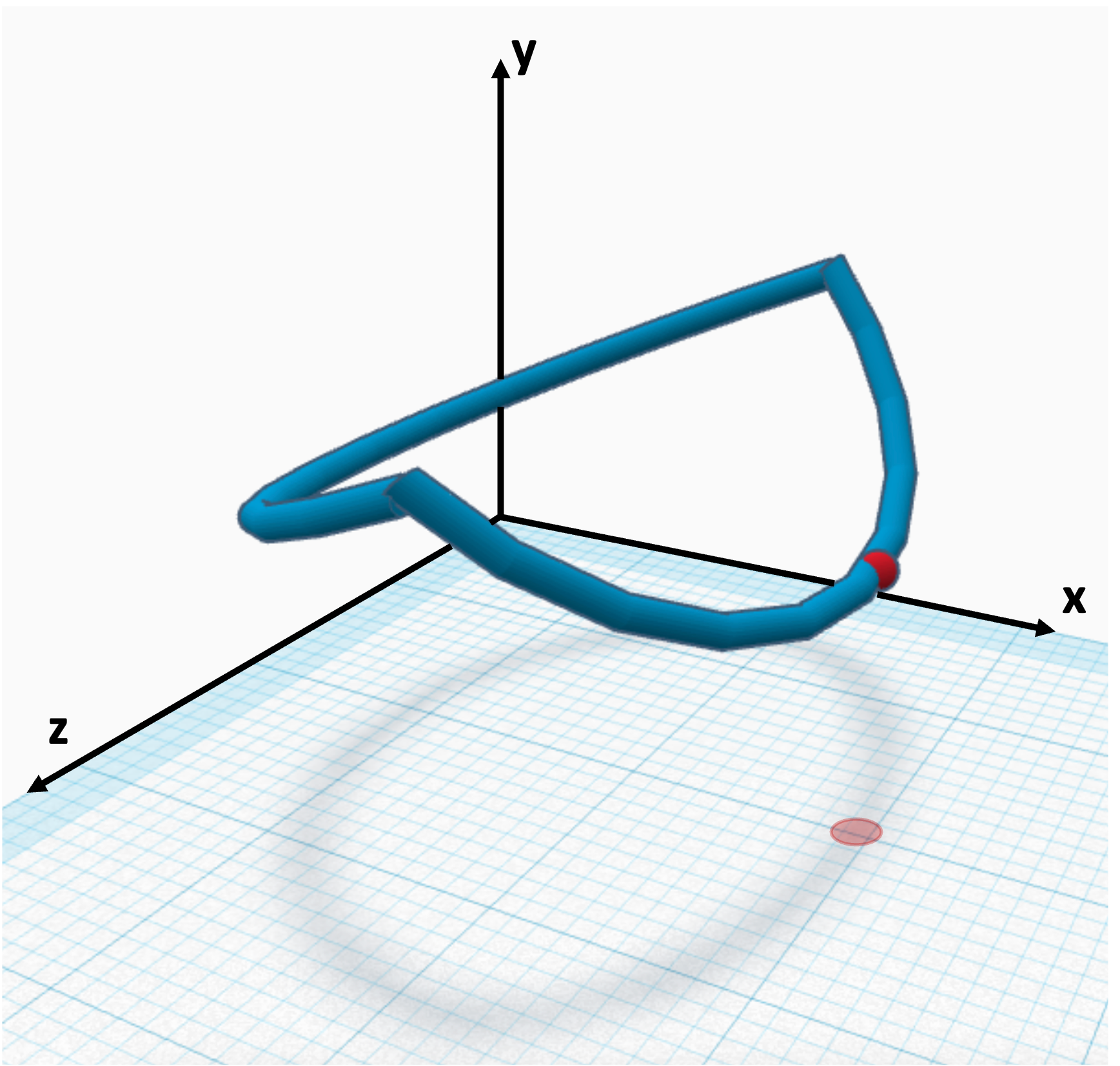} }}%
    \caption{\textit{\textbf{Constrained Intervention}} involves altering the predicted class of the test sample with as few one-dimensional moves as possible. Considering the blue volume as the \textit{normal} region and its surface as the decision hyperplane, (a and b) the test sample has two of the three feature values within the range of the decision hyperplane. (c) Hence, a move along y-axis is sufficient to place the sample within the \textit{normal} zone.}
    \label{i_view}%
\end{figure*}
\vspace{-2mm}
\textit{\textbf{Constrained Intervention:}} To ensure user preferences or adherence to physician recommendations, an additional constraint is needed into the previous optimization framework. While meeting minimal changes to toggle the predicted class, this constraint aims to preserve specific aspects or features of the test sample whenever deemed feasible. Therefore, the minimal adversarial perturbation-
\begin{align*}
\delta_{min,p} &= \min_{\delta \in \mathbb{R}^d}||\delta||p\\
\text{s.t. } &f_n(X_T +\delta) > \max_{a\neq n, \forall a} f_a(X_T +\delta), \text{ } X_T +\delta\in C, \\
&z_i \cdot \delta_i = 0, \text{ } \forall i \text{ where } z_i = \{0,1\} \text{ and } \sum_{i=1}^{d} z_i < d
\end{align*}
\vspace{-0.5mm}
 The key motivation behind this optimization is driven by the fact that if certain features of the test sample fall within the upper and lower bound of the decision boundary, at least one of them can be kept unchanged while changing the predicted class. For example, two out three features of the red test sample in Figure~\ref{ia} and \ref{ib} reside within the range of the blue decision boundary. A mere shift along y-axis is all it takes to change the class of the test sample, while the remaining feature values stay unaffected. Algorithm~\ref{int_alg} executes this subset-based intervention, ensuring that changes are only made to features that are not in their desired position.

Because Algorithm~\ref{int_alg} loops over all the borderline instances to obtain the closest instance, once again, the complexity is $\mathcal{O}(\|S\|+\|\mathcal{P}\|) \rightarrow \mathcal{O}(\|S\|)$, where $\|S\|$ is the number of borderline instances and $\|P\|=d$.
\begin{algorithm}[!h]
\footnotesize
\caption{Subset-based intervention}
\label{int_alg}
\textbf{Input}: Test sample $X_T = {x_T}^1,...,{x_T}^d$, order\footnotemark of features $\mathcal{P}$, classifier $f$, set of borderline instances $S$ \\
\textbf{Output}: Intervened test sample ${X_T}^*$ \\
\textbf{Initialization}: ${X_T}^* \leftarrow \emptyset$
\begin{algorithmic}[1] %[1] enables line numbers
\IF{${x_T}^{\mathcal{P}[1]} > \max(x^{\mathcal{P}[1]} : x \in S)$}
\STATE${x_T}^{\mathcal{P}[1]*} \leftarrow \max(x^{\mathcal{P}[1]} : x \in S)$
\ELSIF{${x_T}^{\mathcal{P}[1]} < \min(x^{\mathcal{P}[1]} : x \in S)$}
\STATE${x_T}^{\mathcal{P}[1]*} \leftarrow \min(x^{\mathcal{P}[1]} : x \in S)$
\ENDIF
\STATE closest index, $\mathcal{C}$ = $\arg\min_{i=1}^{|S|} |S[i]_{\mathcal{P}[1]} - x_T^{\mathcal{P}[1]}|$
\FOR{$i = 2$ to $|\mathcal{P}|$}
\STATE ${x_T}^{\mathcal{P}[i]*} \leftarrow S[i]_{\mathcal{P}
[\mathcal{C}]}$
\IF{$f_n({X_T}^*)>f_a({X_T}^*)$}
\STATE break
\ENDIF
\ENDFOR
\end{algorithmic}
\end{algorithm}

\section{Experimental Setup}
\underline{\textbf{Baselines}} Consistent with prior research \cite{mothilal2020dice, Brughmans2021NICEAA, Mazzine2021AFA, Pawelczyk2021CARLAAP}, our experiments are based on tabular datasets with binary target variables. We compare ExAct against four state-of-the-art CF explanation methods:
%\begin{tasks}[label=\textbullet]

    \textbf{DiCE} \cite{mothilal2020dice} generates a set of CFs by optimizing for proximity, diversity and sparsity. 
    
    \textbf{Optbinning} \cite{Navas-Palencia2021Counterfactual} creates CFs by optimizing a set of binning rules to modify the input features with an aim to find the shortest path to a desired outcome.
    
    \textbf{CEML} \cite{ceml}, like \cite{Schut2021GeneratingIC}, calculates gradients of the output with respect to the inputs and iteratively finds the smallest possible changes to the important features to alter the model's predictions.
     
    \textbf{NICE} \cite{Brughmans2021NICEAA} generates CFs by searching for nearby instances in the data manifold that reflects the desired outcome.
    
%\end{tasks}
\footnotetext{higher indexed features are preferred to remain unchanged.}
Like ExAct, all the baselines are post-hoc methods and require a trained predictive model to proceed. For each dataset, we ensured a fair comparison by employing either the model from ExAct or a model with comparable performance. All train/test splits are done randomly.

\underline{\textbf{Datasets}} ExAct has been tested on four real-world binary classification datasets for chronic disease prevention and management:
\begin{tasks}[label=\textbullet]
    \task Modified \textbf{Heart Disease} \cite{misc_heart_disease_45} from \cite{fedesoriano2021heart} has $918$ instances. We used four categorical and five continuous features to predict whether an individual will have a heart disease $(y=1)$ or not $(y=0)$ and make interventions based on five features.
    \task \textbf{PimaDM} \cite{Smith1988UsingTA} has one categorical and seven continuous attributes from $768$ subjects. The goal is to predict if a subject would have diabetes $(y=1)$ or not $(y=0)$ and make interventions based on four attributes to prevent a positive case.
    \task \textbf{Nutrition Absorption (NA)} \cite{Karim2020AftermealBG} dataset includes $668$ hours' worth of continuous glucose monitor (CGM) recordings resulting from $167$ meals consumed by five diabetes patients. Given a meal, the task is to predict whether a patient would be hyperglycemic $(y=1)$ or remain normoglycemic $(y=0)$ \cite{ElSayed20226GT} and make suggestions on a subset of seven continuous features to prevent a hyperglycemic outcome.
    \task The \textbf{OhioT1DM} \cite{Marling2020TheOD} has eight-week long clinical data including $1600$ hours of CGM signals from $12$ T1 diabetes patients. We predict post-prandial hyperglycemia $(y=1)$ or normoglycemia $(y=0)$ \cite{ElSayed20226GT} and make interventions using a subset of seven continuous features to avoid hyperglycemia.
\end{tasks}

\textbf{\underline{Evaluation Metrics}} Evaluating the explanations of XAI has been a two-decade long struggle \cite{Miller2017ExplanationIA,DoshiVelez2017TowardsAR}. Therefore, we evaluate the CF explanations based on five common metrics found in the literature-

\textbf{Validity} evaluates whether or not the produced CFs really belong to the desired class \cite{Guo2021CounterNetET, Guidotti2022, Upadhyay2021TowardsRA,Jeanneret2023AdversarialCV}. High validity indicates the technique’s effectiveness in creating valid CF examples. Following \cite{Barbiero2021EntropybasedLE}, a simulation-aided method (see Supplementary) is designed to estimate the validity of the CFs.

\textbf{Proximity} is determined by calculating the $L_2$ norm distance between $X_T$ and ${X_T}^*$, and then dividing it by the cardinality of features. \textit{Proximity} helps to ensure we are making minimal change to the factual sample by preserving as much as possible and not over-correcting to hypoglycemia \cite{Wachter2017CounterfactualEW, mothilal2020dice, Mahajan2019PreservingCC, Guo2021CounterNetET}.

\textbf{Sparsity} quantifies the feature changes (via the $L_0$ norm) between $X_T$ and ${X_T}^*$. This comes from the notion of sparse explanations, where fewer feature modifications in CF explanations enhance user interpretability \cite{Wachter2017CounterfactualEW, Miller2017ExplanationIA, PoursabziSangdeh2018ManipulatingAM}. Lower sparsity is preferred for better user understanding \cite{Guo2021CounterNetET}. 

\textbf{Violations} estimates the extent to which user preferences (e.g. age, gender, insulin etc.) are breached. A good CF technique will have fewer violations per generated CF and promote fairness \cite{Wang2019EqualOA, Chen2024CounterfactualFT}. %refers to the average number of feature preferences not preserved in the generated CFs.

\textbf{Plausibility} measures what fraction of the generated CFs fall within the original data manifold \cite{Guidotti2022, Ser2022ExploringTT}. A higher plausibility ensures that the suggested modifications are not difficult-to-impossible to achieve and follow the data distribution \cite{Keane2020GoodCA}.

\section{Results}
\textbf{\underline{Classifier Performance}}
The classifiers trained on the Heart Disease, PimaDM, Nutrition Absorption and OhioT1DM dataset have $81.52\%$, $80.86\%$, $82.4\%$ and $81\%$ prediction accuracy, respectively. Using accuracy metric is logical since the datasets are either originally class-balanced or have been balanced by upsampling.

\begin{figure*}[!h]
    \centering
    \subfloat[\label{v1}]{{\includegraphics[scale=0.33]{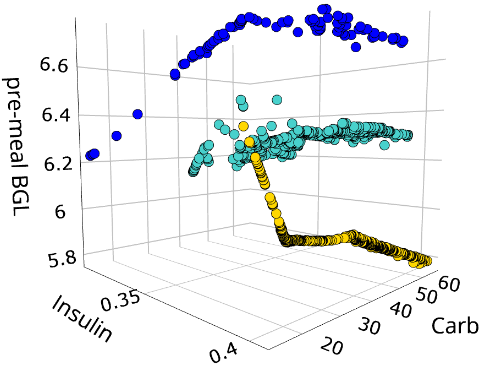} }}%
    \quad
    \subfloat[\label{v5}]{{\includegraphics[scale=0.30]{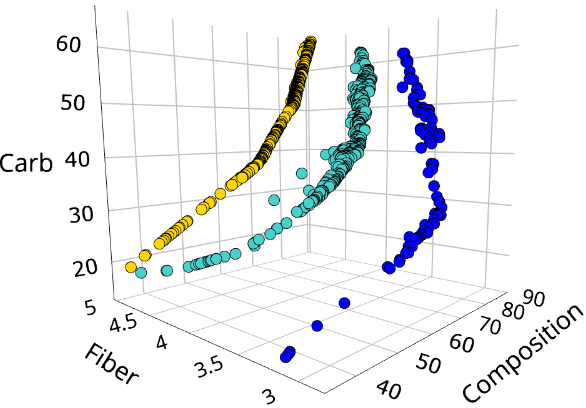} }}%
    \quad
    \subfloat[\label{v3}]{{\includegraphics[scale=0.27,trim=1mm 1mm 1mm 1mm, clip]{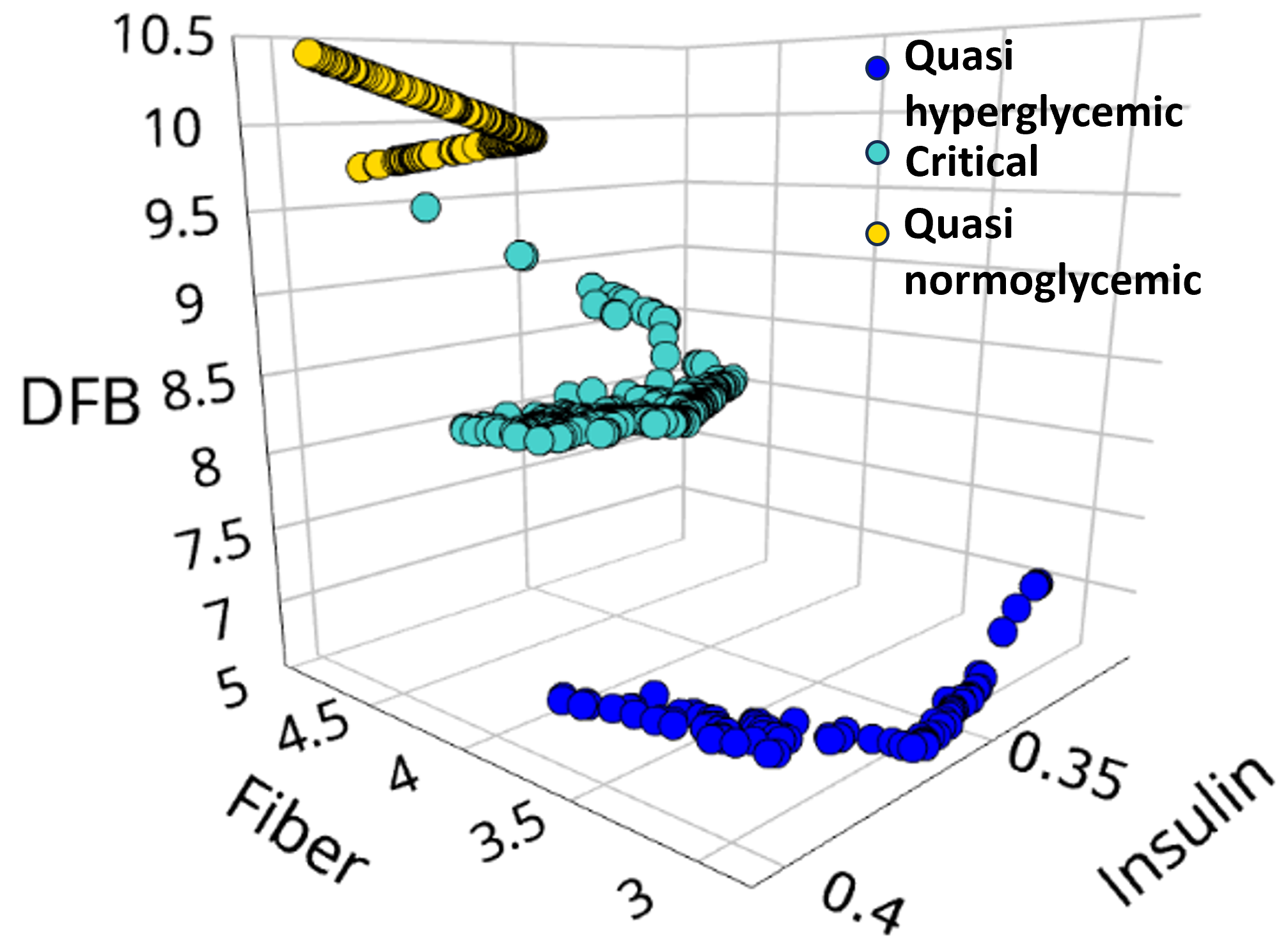} }}%
    \caption{Outputs of the autoencoders and the bisection algorithm, i.e., the borderline instances, are shown across three plots. The plots display six out of the seven features, grouped as (a) pre-meal BGL (i.e. SBGL), insulin intake, CHO, (b) CHO, fiber intake, CHO composition, and (c) DFB (i.e. time elapsed since last insulin), fiber intake, insulin intake. This visualization demonstrates the precise placement of critical instances between the quasi-hyperglycemic and normoglycemic samples.}
    \label{3D}%
\end{figure*}

\textbf{\underline{Autoencoder Performance}}
The effectiveness of ExAct is highly reliant on generating a sufficient number of borderline instances along the decision hyperplane. Higher density of the critical samples on the decision hyperplane is crucial to quality CFs. To achieve a high density of critical samples, the NA dataset has been augmented with 1000 synthetic samples, which are later fed into the autoencoders. This approach allows for a detailed characterization of the decision boundary, resulting in granularity in critical samples and thus, meaningful CFs. Figure~\ref{3D} shows the critical instances along different axes as generated by the autoencoders.

\textbf{\underline{Evaluating the Explanations}} The summary in Table \ref{evaluation} outlines the quality of CFs generated by different methods. ExAct achieves an average validity that surpasses DiCE and CEML by $11\%$ and exceeds Optbinning and NICE by $15\%$. Note that, no technique gets a perfect score as the CFs are evaluated by external simulators and not by the corresponding classifiers. The CFs produced by ExAct are at least $6.6\%$ closer to the test samples than those from other techniques on three (Heart Disease, PimaDM and OhioT1DM) out of four datasets. 

ExAct leads the list in preserving user preferences with the least number of feature violations per explanation. It achieves perfect violation scores for two datasets and incurs a few violations in the other two. With 1.68 feature changes per CF explanation on average, ExAct is only behind DiCE in terms of sparsity and leads the other three methods in most datasets.

\begin{wrapfigure}{r}{6.4cm}
\vspace{-5mm}
\includegraphics[width=0.9\linewidth]{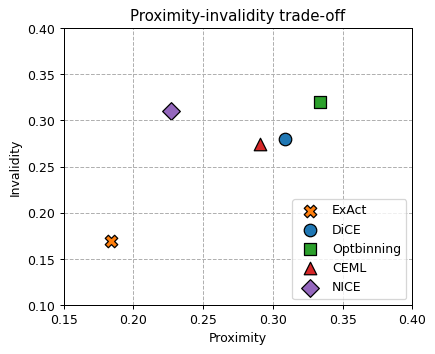}
\caption{Proximity-invalidity trade-off across all methods.} 
\label{tradeoff}
\end{wrapfigure}

ExAct falls behind NICE on two datasets (Nutrition Absorption and OhioT1DM) in generating plausible explanations. However, NICE's ability to maintain $100\%$ plausibility stems from the fact that it sources CF examples directly from the dataset without generating any new samples. Based on the two intervention plans, a pair of explanations are listed in Table~\ref{explanations}. Figure~\ref{tradeoff} shows proximity-invalidity trade-off for all methods.

\textbf{\underline{Limitations}} One limitation is that ExAct, like other approaches \cite{mothilal2020dice, Karimi2019ModelAgnosticCE, Navas-Palencia2021Counterfactual, Poyiadzi2019FACEFA}, may incur high computational overhead in a multi-class setting.
\begin{table*}[t]
\footnotesize
\caption{Evaluating the CFs: ExAct outperforms others in validity, proximity, violations and achieves comparable results in sparsity and plausibility.}
\label{evaluation}
\centering
\begin{tabular}{p{0.6in}|C{0.28in}C{0.28in}C{0.28in}C{0.28in}C{0.28in}|C{0.28in}C{0.28in}C{0.28in}C{0.28in}C{0.28in}}
\toprule
 \multirow{2}{*}{Method} &\multicolumn{5}{c|}{\textbf{Heart Disease}}& \multicolumn{5}{c}{\textbf{PimaDM}} \\
& \cellcolor{blue!18}val. & \cellcolor{red!25}prox. & \cellcolor{green!25}spar. & \cellcolor{yellow!25}viol. & \cellcolor{teal!25}plau. & \cellcolor{blue!18}val. & \cellcolor{red!25}prox. & \cellcolor{green!25}spar. & \cellcolor{yellow!25}viol. & \cellcolor{teal!25}plau. \\
\midrule 
ExAct & \textbf{0.85} & \textbf{0.083} & 1.8 & \textbf{0} &  \textbf{1.0} & \textbf{0.74} & \textbf{0.099} & 1.66 & \textbf{0} &  \textbf{1.0} \\
DiCE & 0.73 & 0.262 & \textbf{1.7} & \textbf{0} &  \textbf{1.0} & 0.63 & 0.149 & \textbf{1.29} & \textbf{0} & \textbf{1.0} \\
Optbinning & \textbf{0.85} & 0.302 & 5.1 & 2.6 & \textbf{1.0} & 0.61 & 0.361 & 2.29 & \textbf{0} &  \textbf{1.0} \\
CEML & 0.78 & 0.274 & 2.95 & 2.8 &  \textbf{1.0} & 0.6 & 0.138 & 1.5 & 0.5 & \textbf{1.0} \\
NICE & 0.83 & 0.088 & 2.3 & 0.3 &  \textbf{1.0} & 0.71 & 0.13 & 1.84 & 0.32 &  \textbf{1.0} \\
\midrule \midrule
&\multicolumn{5}{c|}{\textbf{Nutrition Absorption}}& \multicolumn{5}{c}{\textbf{OhioT1DM}} \\
& \cellcolor{blue!18}val. & \cellcolor{red!25}prox. & \cellcolor{green!25}spar. & \cellcolor{yellow!25}viol. & \cellcolor{teal!25}plau. & \cellcolor{blue!18}val. & \cellcolor{red!25}prox. & \cellcolor{green!25}spar. & \cellcolor{yellow!25}viol. & \cellcolor{teal!25}plau. \\
\midrule 
ExAct & \textbf{0.83} & 0.239 & \textbf{1.33} & \textbf{0.05} &  0.98 & \textbf{0.9} & \textbf{0.314} & \textbf{1.93} & \textbf{0.05} & 0.9 \\
DiCE & 0.73 & 0.353 & 1.35 & 0.1 &  0.9 & 0.8 & 0.472 & \textbf{1.93} & \textbf{0.05} & 0.83\\
Optbinning & 0.7 & 0.287 & 1.78 & 0.15 &  0.95 & 0.75 & 0.387 & 2.1 & 0.18 &  0.88 \\
CEML & 0.75 & 0.314 & 1.48 & 0.15 & 0.95 & 0.78 & 0.438 & 2.55 & 0.18 & 0.9 \\
NICE & 0.43 & \textbf{0.176} & 2.2 & 0.23 & \textbf{1.0} & 0.8 & 0.518 & 2.95 & 0.3 & \textbf{1.0} \\
\bottomrule
\end{tabular}
\end{table*}

\begin{table*}[!h]
\caption{Examples of minimal and constrained interventions made for glucose control.}
\label{explanations}
\centering
\footnotesize
\begin{tabular}{p{3.3in}|p{1.8in}}
\toprule
\multicolumn{2}{c}{\textbf{Nutrition Absorption dataset}}  \\
\midrule
\textbf{Pre-meal context}& \textbf{Intervention} \\
\midrule
Aaron's current blood glucose level is 6.8 mmol/L. His Lunch contains 60.5g CHO, 9.4g fat and 4.2g fiber. He took 0.32 unit bolus 8 minutes before lunch. He is predicted to experience post-prandial hyperglycemia. & (\textbf{minimal}) Increase CHO by 1.3g, fat by 2.6g, take a bolus of 0.4 units and wait until blood glucose drops to 6 mmol/L to stay normoglycemic. \\ \midrule
Emily prepared a meal with 74.2g CHO, 6.4g fat and 3.1g fiber. She has already taken 0.32 unit bolus 5 minutes back and her glucose level is 7.2 mmol/L. Emily wants to eat her meal while avoiding another insulin dose to stay normoglycemic. & (\textbf{constrained}) She can remain normoglycemic just by reducing the carb amount to 21.3 grams. \\
\bottomrule
\end{tabular}
\end{table*}

\section{CONCLUSION}
In this paper, ExAct is proposed as a technique for generating counterfactual explanations to enhance model interpretability and ensure trust and fairness in AI. Unlike many prior works, ExAct promises to reflect user preferences in the explanations by keeping certain features unchanged. Being model-agnostic, ExAct approximates the decision boundary of the classifier using critical adversarial instances and conducts a sophisticated grid search to ensure user preferences are preserved. Analysis conducted on four datasets demonstrates that ExAct outperforms other methods in terms of validity, proximity, and violation metrics, while also exhibiting competitive performance in terms of sparsity and plausibility. In summary, the proposed approach represents a significant milestone in the field of generative counterfactual explanations.

%%%%%%%%%%%%%%%%%%%%%%%%%%%%%%%%%%%%%%%%%%%%%%%%%%%%%%%%%%%%

\newpage
%\bibliographystyle{unsrt}

%%%%%%%%%%%%%%%%%%%%%%%%%%%%%%%%%%%%%%%%%%%%%%%%%%%%%%%%%%%%

\newpage
\begin{center}
    \part*{Technical Appendix}
    \addcontentsline{toc}{part}{Technical Appendix}
\end{center}
\setcounter{section}{0}

\section{Resource Specifications}
All experiments are done with an AMD Ryzen 7 2700X 8-core CPU of 3.7 GHz speed, an NVIDIA GeForce GTX 1660 Ti GPU and 16 gigabyte RAM.

\section{Dataset Descriptions}

\subsection{Heart Disease}
The dataset contains patient characteristics and test results to predict the presence or absence of heart disease. The attributes used in training include Age, Sex, Chest pain type, Resting BP, Cholesterol, Fasting blood sugar, Resting electrocardiographic results, Maximum heart rate, ST depression induced by exercise. The explanations are based on changing Resting BP, Cholesterol, Fasting blood sugar, Maximum heart rate and ST depression.

\subsection{PimaDM}
The \textbf{PimaDM} dataset is a widely used benchmark dataset for predicting diabetes based on patient characteristics. It contains contains medical records of $768$ Pima Indian women aged $21$ and above, living in Phoenix, Arizona. The prediction model is trained using all eight features (number of pregnancies, glucose level, blood pressure, skin, thickness, insulin, BMI, diabetes pedigree function and age) while we changed only four features (glucose level, blood pressure, insulin and BMI) in the explanations and other features are preferred to be kept unchanged. The goal is to predict whether a subject will have diabetes or not given the features. There are 268 positive cases and 500 negative cases. We carefully up-sampled the positive cases randomly in the training data and balanced it.

\subsection{Nutrition Absorption Dataset}
The \textbf{Nutrition Absorption} dataset includes CGM recordings from 5 participants (4 male, 1 female, Age: 51.6 $\pm$ 16.1 y). All individuals were diabetic patients and given a total of 167 meals under fasting condition. The dataset includes meal macronutrient (carbohydrates, fat, and fiber) amounts, time elapsed since the last insulin dose (DFB), pre-meal blood glucose levels (SBGL), and CGM data spanning four hours post-meal.  Since this dataset is small, we supplemented it with synthetic data made with conditional autoregressive model [1]. More details can be found in the FAQs. 

\subsection{OhioT1DM}
The \textbf{OhioT1DM} is an eight-week long clinical study of 12 deidentified T1 diabetes patients. Participants used insulin pumps and CGM sensors. Their physiological data (acceleration, skin response etc.) was recorded on wristbands while their self-reported CHO intake, work and exercise intensities were recorded on smartphones. Only seven subjects have been included in analysis to ensure coherence in input data.

The input to the prediction model is Carbohydrate intake, total basal intake, bolus intake, time since last bolus, total exercise (intensity $\times$ duration), total work (intensity $\times$ duration) and pre-meal blood glucose level from the past three-hour. The task is to predict post-prandial hyperglycemia for a two-hour long window. Each post-prandial glycemic response (PPGR) was categorized into hyperglycemia (\textit{max}(PPGR) $\geq$ 181 mmol/L) or normoglycemia (\textit{max}(PPGR) $<$ 181 mmol/L) as target classes.

\section{Model Specifications}
Model hyperparameters are tuned on a trial-and-error basis. All results are generated with the best set of hyper-parameters identified.

\subsection{Heart Disease}

\subsubsection{Classifier}

The fully-connected binary classifier is described in Table~\ref{tab:classifier_HD}-

\begin{table}[ht]
\footnotesize
\centering
\caption{Classifier Specifications for Heart Disease data}
\label{tab:classifier_HD}
\begin{tabular}{>{\raggedright\arraybackslash}p{2.6cm} p{8.5cm}}
\toprule
\textbf{Layer} & \textbf{Description} \\
\midrule
\textbf{Input Layer} & Dense, 16 neurons, \textit{ELU} activation, \textit{$l_2$} regularizer (factor: 0.04), \textit{HeNormal} initializer, Batch normalization, Dropout rate: 0.2 \\
\midrule
\textbf{Hidden Layer 1} & Dense, 8 neurons, \textit{ELU} activation, \textit{$l_2$} regularizer (factor: 0.04), \textit{HeNormal} initializer, Batch normalization, Dropout rate: 0.2 \\
\midrule
\textbf{Output Layer} & Dense, 2 neurons, \textit{Sigmoid} activation \\
\midrule
\textbf{Optimizer} & Adam, learning rate: 0.001 \\
\midrule
\textbf{Dataset Split} & 70\% training set, 30\% test set \\
\midrule
\textbf{Training} & 100 epochs, batch size: 16 \\
\bottomrule
\end{tabular}
\end{table}

\subsubsection{Autoencoders}
Table~\ref{tab:autoencoder_HD} describes the indentical autoencoders used for critical sample generation.

\begin{table}[ht]
\footnotesize
\centering
\caption{Autoencoder Specifications for Heart Disease dataset}
\label{tab:autoencoder_HD}
\begin{tabular}{>{\raggedright\arraybackslash}p{4cm} p{8cm}}
\toprule
\textbf{Layer} & \textbf{Description} \\
\midrule
\textbf{Encoder} & \\
Layer 1 & Dense, 16 neurons, \textit{ReLU} activation \\
Layer 2 & Dense, 8 neurons, \textit{ReLU} activation \\
\midrule
\textbf{Code} & \\
Code Layer & Dense, 8 neurons, \textit{ReLU} activation \\
\midrule
\textbf{Decoder} & \\
Decoding Layer 1 & Dense, 16 neurons, \textit{ReLU} activation \\
Output (Gender) & Dense, 2 neurons, \textit{Softmax} activation \\
Output (CPT) & Dense, 4 neurons, \textit{Softmax} activation \\
Output (FBGL) & Dense, 2 neurons, \textit{Softmax} activation\\
Output (RestECG) & Dense, 2 neurons, \textit{Softmax} activation\\
Output (Continuous) & Dense, 5 neurons, \textit{ReLU} activation\\
Concatenate & Combine all output branches \\
\midrule
\textbf{Optimizer} & Adam, learning rate: 0.001 \\
\textbf{Loss Function} & Custom loss where $\alpha = 80$\\
\midrule
\textbf{Training} & 120 epochs, batch size: 16 \\
\bottomrule
\end{tabular}
\end{table}

\subsection{PimaDM}

\subsubsection{Classifier}
The fully-connected binary classifier is described in Table~\ref{tab:clf_PimaDM}-

\begin{table}[htbp]
\footnotesize
\centering
\caption{Classifier Specifications for PimaDM}
\label{tab:clf_PimaDM}
\begin{tabular}{>{\raggedright\arraybackslash}p{3.5cm} p{8cm}}
\toprule
\textbf{Layer} & \textbf{Description} \\
\midrule
\textbf{Input Layer} & Dense, 64 neurons, \textit{LeakyReLU} activation, HeNormal initializer, Batch Normalization, Dropout rate: 0.4  \\
\midrule
\textbf{Hidden Layer 1} & Dense, 32 neurons, \textit{LeakyReLU} activation, HeNormal initializer, Batch Normalization, Dropout rate: 0.4 \\
\midrule
\textbf{Hidden Layer 2} & Dense, 16 neurons, \textit{LeakyReLU} activation, HeNormal initializer, Batch Normalization, Dropout rate: 0.2 \\
\midrule
\textbf{Output Layer} & Dense, 2 neurons, \textit{sigmoid} activation \\
\midrule
\textbf{Optimizer} & Adam, learning rate: 0.01 \\
\midrule
\textbf{Dataset Split} & 80\% training set, 20\% test set \\
\midrule
\textbf{Training} & 80 epochs, batch size: 16 \\
\bottomrule
\end{tabular}
\end{table}

\subsubsection{Autoencoders}
Specifications of the two identical autoencoders are given in Table~\ref{tab:autoencoder_PimaDM}.

\begin{table}[htbp]
\footnotesize
\centering
\caption{Autoencoder Specifications for PimaDM}
\label{tab:autoencoder_PimaDM}
\begin{tabular}{>{\raggedright\arraybackslash}p{4cm} p{8cm}}
\toprule
\textbf{Layer} & \textbf{Description} \\
\midrule
\textbf{Encoder} & \\
Layer 1 & Dense, 64 neurons, \textit{ReLU} activation \\
Dropout 1 & Rate: 0.1 \\
Layer 2 & Dense, 32 neurons, \textit{ReLU} activation \\
Dropout 2 & Rate: 0.1 \\
\midrule
\textbf{Code} & \\
Code Layer & Dense, 16 neurons, \textit{ReLU} activation \\
\midrule
\textbf{Decoder} & \\
Layer 1 & Dense, 32 neurons, \textit{ReLU} activation \\
Dropout 3 & Rate: 0.1 \\
Layer 2 & Dense, 64 neurons, \textit{ReLU} activation \\
Dropout 4 & Rate: 0.1 \\
Output Layer & Dense, $X_1.shape[1]$ neurons, \textit{ReLU} activation \\
\midrule
\textbf{Optimizer} & Adam, learning rate: 0.001 \\
\textbf{Loss Function} & Custom loss where $\alpha = 20$\\
\midrule
\textbf{Training} & 50 epochs, batch size: 16 \\
\bottomrule
\end{tabular}
\end{table}

\subsection{Nutrition Absorption Data}
\subsubsection{Classifier}
The fully-connected binary classifier is described in Table~\ref{tab:classifier_NA}-

\begin{table}[ht]
\footnotesize
\centering
\caption{Classifier Specifications for Nutrition Absorption}
\label{tab:classifier_NA}
\begin{tabular}{>{\raggedright\arraybackslash}p{4cm} p{8cm}}
\toprule
\textbf{Parameter} & \textbf{Description} \\
\midrule
\textbf{Model Architecture} & Fully-connected network (two layers) \\
\textbf{Neurons per Layer} & 32 \\
\textbf{Activation Function} & LeakyReLU \\
\textbf{Regularization} & $l_2$ regularizer with factor $0.04$ \\
\textbf{Dropout Rate} & 0.5 \\
\textbf{Optimizer} & SGD, learning rate: $1e-4$, Weight Decay: $1e-5$ \\
\textbf{Dataset Split} & 80\% training set, 20\% test set \\
\textbf{Synthetic Samples} & Approximately 400 used for model training \\
\bottomrule
\end{tabular}
\end{table}
\subsubsection{Autoencoders}
Two identical fully-connected autoencoders have been fed with hyperglycemic and normoglycemic instances respectively to create critical samples. For any hyperglycemic sample, the nearest normoglycemic from the data is set as the label. This process is reversed for normoglycemic samples. More details are given in Table~\ref{tab:autoencoders_NA}-

\begin{table}[ht]
\footnotesize
\centering
\caption{Autoencoder Specifications for Nutrition Absorption}
\label{tab:autoencoders_NA}
\begin{tabular}{>{\raggedright\arraybackslash}p{3.5cm} p{8.5cm}}
\toprule
\textbf{Parameter} & \textbf{Description} \\
\midrule
\textbf{Data Input} & Hyperglycemic and normoglycemic instances \\
\textbf{Code Layer Neurons} & 8 \\
\textbf{Encoder/Decoder} & 2 layers with 16 neurons each \\
\textbf{Activation Function} & LeakyReLU \\
\textbf{Regularization} & $l_2$ regularizer with factor $0.02$ \\
\textbf{Dropout Rate} & 0.5 \\
\textbf{Optimizer} & Adam \\
\textbf{Learning Rate} & $1e-3$ \\
\textbf{Weight Decay} & $1e-5$ \\
\textbf{$\alpha$ value} & $\alpha = 4$ in Equation 3 for evenly distributed samples, prioritizing cross-entropy loss \\
\textbf{Alpha Adjustment} & Lower $\alpha$ values prioritize reconstruction loss, concentrating points around a single feature space point \\
\bottomrule
\end{tabular}
\end{table}

\subsection{OhioT1DM}
\footnotesize
\subsubsection{Classifier}
The binary classifier for OhioT1DM is described in Table~\ref{tab:classifier_Ohio}.

\begin{table}[h]
\centering
\caption{Classifier Specifications for OhioT1DM}
\label{tab:classifier_Ohio}
\begin{tabular}{>{\raggedright\arraybackslash}p{3.5cm} p{8cm}}
\toprule
\textbf{Parameter} & \textbf{Description} \\
\midrule
\textbf{Model Architecture} & Fully-connected network (four layers) \\
\textbf{Neurons per Layer} & 256, 128, 64, and 32 neurons, respectively \\
\textbf{Activation Function} & \textit{tanh} \\
\textbf{Regularization} & $l_2$ regularizer with factor $0.01$ \\
\textbf{Dropout Rate} & 0.4 \\
\textbf{Optimizer} & Adam \\
\textbf{Learning Rate} & $1e-4$ \\
\textbf{Weight Decay} & $1e-5$ \\
\bottomrule
\end{tabular}
\end{table}

\subsubsection{Autoencoders}
The identical autoencoders for the OhioT1DM is described in Table~\ref{tab:autoencoders_Ohio}

\begin{table}[ht]
\footnotesize
\centering
\caption{Autoencoder Specifications for OhioT1DM}
\label{tab:autoencoders_Ohio}
\begin{tabular}{>{\raggedright\arraybackslash}p{3.5cm} p{8cm}}
\toprule
\textbf{Parameter} & \textbf{Description} \\
\midrule
\textbf{Layers} & Two fully connected layers for encoder and decoder \\
\textbf{Neurons per Layer} & Encoder/Decoder: 150, 64 neurons \\
\textbf{Code Layer} & 32 neurons, \textit{ReLU} activation \\
\textbf{Activation Function} & \textit{tanh} for encoder and decoder \\
\textbf{Regularization} & $l_2$ regularizer with factor $0.02$ \\
\textbf{Dropout Rate} & 0.1 \\
\textbf{Optimizer} & Adam \\
\textbf{Learning Rate} & $1e-3$ \\
\textbf{Weight Decay} & $1e-5$ \\
\textbf{$\alpha$ value} & $\alpha = 3$ in Equation 3 for diversified samples \\
\bottomrule
\end{tabular}
\end{table}

\section{Diversity}
\begin{figure}[!h]
  \centering
  \begin{subfigure}[b]{0.48\textwidth} % Subfigure 1
    \centering
    \includegraphics[width=\textwidth]{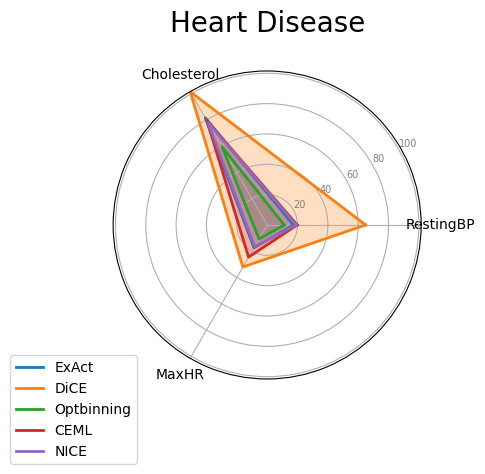}
    \label{HD_d}
  \end{subfigure}
  \hfill
  \begin{subfigure}[b]{0.49\textwidth} % Subfigure 2
    \centering
    \includegraphics[width=\textwidth]{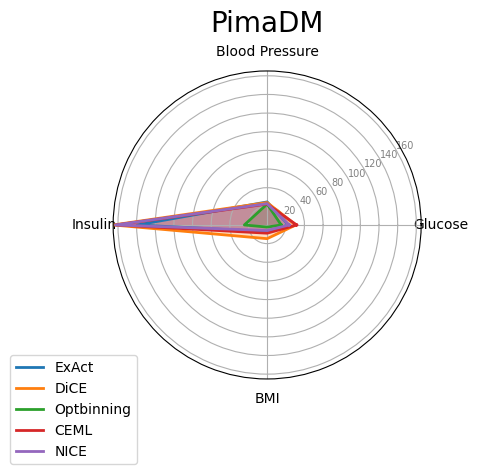}
    \label{PimaDM_d}
  \end{subfigure}
  \caption{Comparing average feature diversity for different methods in Heart Disease and PimaDM dataset.}
  \label{fig:diversity}
\end{figure}

ExAct does not optimize for diversity. Yet, it shows better average diversity for the continuous features compared to NICE and Optbinning in the Heart Disease dataset, and only NICE in the PimaDM dataset. Figure~\ref{fig:diversity} and show the corresponding average diversities in radar plots.

\section{Evaluation Metrics}
\subsection{Validity}
Since the counterfactual explanations are samples themselves, they are fed into a model (external simulator) to verify whether they indeed belong to the desired class. Based on the simulation results, \textbf{\textit{validity}} represents the fraction of counterfactuals that successfully flip to the desired class.
\begin{equation*}
    \textit{validity} = \frac{\#|f(X_T^*) \neq f(X_T)|}{\|CF\|}
\end{equation*}

\subsection{Proximity}
\textbf{\textit{Proximity}} between the counterfactual and the factual sample is calculated from the $L_2$ normalized distance of the continuous features and the hamming distance of the categorical features:

\begin{equation}
\textit{proximity} = \sqrt{\left( \sqrt{\sum_{i=1}^{m_1} \left( \frac{x_T^{*i}}{\|x_T^{*i}\|_2} - \frac{x_T^{i}}{\|x_T^{i}\|_2} \right)^2} \right)^2 + \left( \frac{1}{m_2} \sum_{j=1}^{m} \mathbb{I}(x_T^{*j} \neq x_T^{j}) \right)^2}
\end{equation}

where:
$m_1$ and $m_2$ refer to the number of continuous and categorical features, respectively, with features denoted by superscript $i$ for continuous and $j$ for categorical features.

\subsection{Sparsity}
\textbf{\textit{Sparsity}} is defined as the average number of feature changes per counterfactual example.
\[
\textit{sparsity} = \frac{\sum_{X_T^*\in CF}^{}\sum_{k=1}^{d} \mathbbm{1}(x_T^{*i} \neq x_T^i)}{\|CF\|}
\]

\subsection{Violations}
\textbf{\textit{Violations}} metric is a measure of the average number of feature changes per counterfactual example that were preferred to remain unchanged.
\[
\textit{violations} = \frac{\sum_{X_T^*\in CF}^{}\sum_{k=1}^{d_{\text{\st{mod}}}} \mathbbm{1}(x_T^{*k} \neq x_T^k)}{\|CF\|}
\]

\subsection{Plausibility}
\textbf{\textit{Plausibility}} captures the fraction of explanations that satisfy certain rules set by the data. In this evaluation method, it is defined as the fraction of counterfactuals that fall within the feature ranges derived from the data-
\begin{equation*}
    \textit{Plausibility}=\frac{\sum_{X_T^*\in CF}^{}\mathbbm{1}(\text{dist}(X_T^*)\subseteq\text{dist}(X))}{\|CF\|}
\end{equation*}
where, dist($X_T^*$) represents the distribution of feature values in the counterfactual instance $X_T^*$, dist($X$) represents the distribution of feature values in the training data and $\|CF\|$ is the total number of counterfactual instances.

\subsection{Feature diversity}
Average feature \textit{diversity} has been calculated using the following formula-
\begin{equation*}
    \text{Average diversity for feature }k = \frac{\sum_i\sum_j|{x_i}^k-{x_j}^k|}{\|CF\|}, i\neq j
\end{equation*}

\section{Simulator Specifications}
\begin{figure}[!h]
  \centering
  \begin{subfigure}[b]{0.48\textwidth} % Subfigure 1
    \centering
    \includegraphics[width=\textwidth]{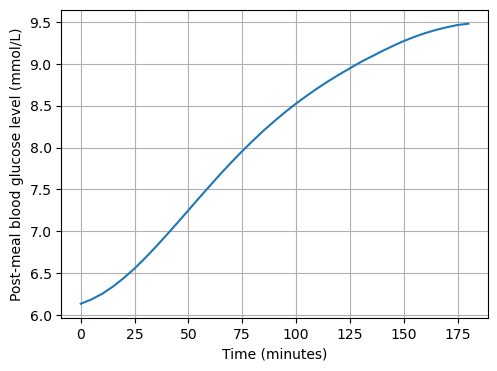}
    \caption{Normoglycemic response of a counterfactual explanation simulated with an external simulator.}
    \label{fig:subfig1}
  \end{subfigure}
  \hfill
  \begin{subfigure}[b]{0.49\textwidth} % Subfigure 2
    \centering
    \includegraphics[width=\textwidth]{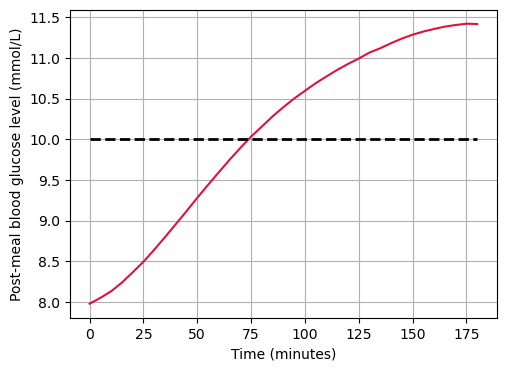}
    \caption{Hyperglycemic response of a counterfactual explanation simulated with an external simulator.}
    \label{fig:subfig2}
  \end{subfigure}
  \caption{Simulated outcomes of counterfactual explanations.}
  \label{fig:entirefig}
\end{figure}

The external simulators for Nutrition Absorption data are subject specific autoregressive models with 4-hour long prediction horizon [anonymous citation available] while the ones for OhioT1DM are subject specific stacked CNN-LSTM regression models that forecast next 3-hour blood glucose levels given meal, insulin, pre-meal blood glucose level, exercise and work information of past hour [anonymous citation available]. The subject specific simulators for OhioT1DM have MAEs ranging from 9.9 mg/dL to 18.8 mg/dL. Figure~\ref{fig:subfig1} and \ref{fig:subfig2} show simulated outcomes of two different cases.

The simulators for the Heart Disease and PimaDM datasets are XGBoost and LGBM models, with accuracies of 85.65\% and 82.78\%, respectively. The simlator specifications are given in Table~\ref{tab:xgboost_params} and \ref{tab:lgbm_params}

\begin{table}[h]
    \centering
    \begin{minipage}{0.45\textwidth}
        \centering
        \caption{Specification of the Simulator for Heart Disease dataset}
        \begin{tabular}{|l|l|}
            \hline
            \textbf{Parameter}         & \textbf{Value}                  \\ \hline
            \hline
            architecture               & XGBoost                         \\ \hline
            n\_estimators              & 600                             \\ \hline
            max\_depth                 & 9                               \\ \hline
            learning\_rate             & 0.0237                          \\ \hline
            subsample                  & 0.2293                          \\ \hline
            colsample\_bytree          & 0.857                           \\ \hline
            gamma                      & 0.0301                          \\ \hline
            min\_child\_weight         & 3.400                           \\ \hline
            accuracy                   & 85.65\%                         \\ \hline
        \end{tabular}
        \label{tab:xgboost_params}
    \end{minipage}%
    \hfill
    \begin{minipage}{0.45\textwidth}
        \centering
        \caption{Specification of the Simulator for PimaDM}
        \begin{tabular}{|l|l|}
            \hline
            \textbf{Parameter}        & \textbf{Value}                    \\ \hline
            \hline
            architecture              & LGBM model                        \\ \hline
            boosting\_type            & dart                              \\ \hline
            objective                 & binary                            \\ \hline
            metric                    & binary\_error                     \\ \hline
            num\_leaves               & 5                                 \\ \hline
            learning\_rate            & 0.1                               \\ \hline
            feature\_fraction         & 0.8                               \\ \hline
            bagging\_fraction         & 0.8                               \\ \hline
            bagging\_freq             & 3                                 \\ \hline
            verbose                   & 0                                 \\ \hline
            accuracy                  & 82.78\%                           \\ \hline
        \end{tabular}
        \label{tab:lgbm_params}
    \end{minipage}
\end{table}

\section{FAQs}
\begin{itemize}
\item What is the time complexity of the grid search to identify the counterfactuals? Does it scale well with larger datasets?

- The \textbf{grid search} algorithm has three key components within it: \textbf{i.} Finding the min and max values of the first feature in the order of features $\mathcal{P}$ from the critical samples $S$, which contributes $O(\|S\|)$ to the complexity. \textbf{ii.} Finding the closest index $\mathcal{C}$ requires iterating over all elements in $S$ to compute the absolute differences and find the minimum. This step also contributes $O(\|S\|)$ to the complexity. \textbf{iii.} The for loop iterates over each element in $\mathcal{P}$, which has a cardinality of $\|\mathcal{P}\|$. Inside the loop, the operations are constant time, so the loop contributes $O(\|\mathcal{P}\|)$ to the complexity.

Hence, the time complexity of the algorithm is $O(\|S\| + \|\mathcal{P}\|)$ or $O(\|S\|)$, where $\|S\|$ is the size of the set of borderline instances, $\|\mathcal{P}\|$ is the number of features and $\|S\|>>\|\mathcal{P}\|$.

In practice, the grid search mostly converged in fraction of a second. With an increased feature number, the grid search might take a bit longer but the complexity will remain $O(|S|)$ as $|S|>>|\mathcal{P}|$. The impact of the number of features on the runtime is likely to be minimal compared to the size of the set of borderline instances.

\item Does the decision boundary approximation pipeline scale well with a larger dataset?

-The \textbf{decision boundary approximation} approach is time consuming as it requires training two autoencoders followed by the bisection method. If number of features increases, the autoencoders will take more time to train as well as the bisection method to converge. However, the runtime can be improved by modifying the condition to converge i.e., bisection will run for a maximum of 20 iterations for each pair etc.

the NA dataset is small and we supplemented it using synthetic data. Thus, the classification accuracy improved from 76\% to 82.4\% and we also got granularity in the approximated decision boundary when we fed them to the autoencoders.

\item The Nutrition Absorption data is small, how it was supplemented using synthetic data?

-The augmentation technique involves using a conditional autoregressive model from the Synthetic Data Vault (SDV) package \cite{Zhang2022SequentialMI} [1] to generate synthetic glycemic response given new context (i.e. feature values). Using the model, we wanted to answer, 'how the glycemic response would be if the subject is fed certain food under certain medication (insulin) and health condition (pre-meal blood glucose level)?'. However, new context is not readily available. So, we pulled macronutrients amounts of food items listed in the USDA database. We observed no significant correlation (P1: -0.05, P2: -0.29, P3: -0.08, P4: 0.02, P5: 0.02) between subjects’ blood glucose levels and their carb (or fat/fiber) intake. As the subjects’ consumed foods were independent of their pre-meal blood glucose level, it is reasonable to place random blood glucose levels (within data range) in each corresponding context. We discovered considerable positive correlation between the insulin intakes and carb intakes (P1: 0.3, P2: 0.76, P3: 0.4, P4: 0.71, P5: 0.21). Therefore, we trained a polynomial regression model using macro-nutrients and insulin amounts from the Absorption dataset and then applied this model to impute insulin intakes. Once we have new context, we fed them to the trained autoregressive models to generate the corresponding glycemic responses. Just to clarify, we did not synthesize data for OhioT1DM. 

\item Why NICE achieved better proximity for Nutrition Absorption data?
    
- Deep learning models are slightly inaccurate on the training data to avoid overfitting. This way, some of the normoglycemic training instances end up on the hyperglycemic side of the classifier. NICE identifies the nearest training sample that has the desired class and returns it as the counterfactual. So, for a factual sample, when the nearest normoglycemic instance is searched from the training data, the nearest normoglycemic sample found could be one from the hyperglycemic side of the classifier which incurs small proximity compared to that of ExAct which searches adversarial normoglycemic samples lying close to the decision boundary. 
\begin{figure*}[!h]
\centering
\includegraphics[width=0.4\linewidth,trim=1mm 1mm 1mm 1mm, clip]{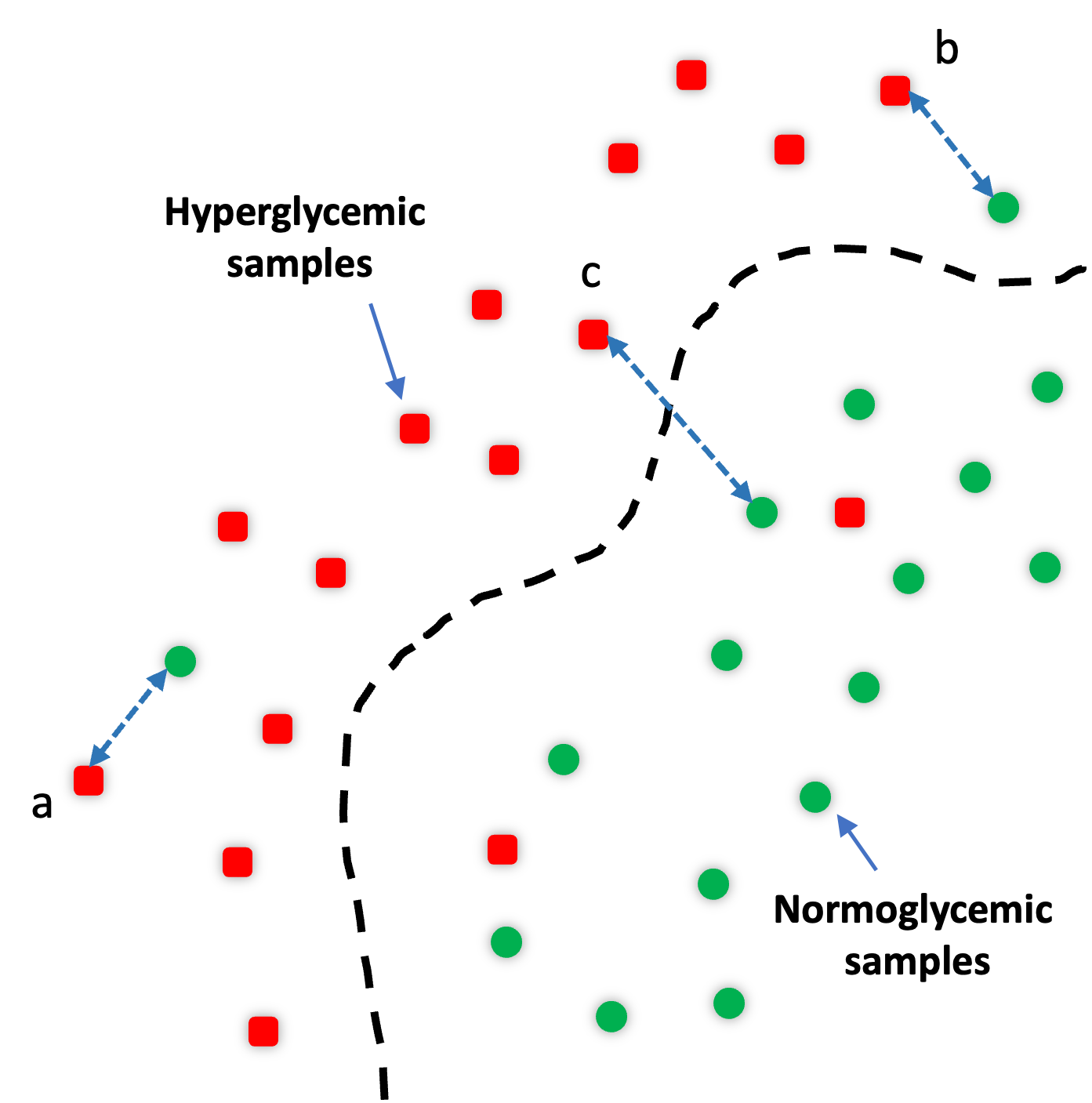}
\caption{As we prevent the classifier from overfitting, NICE achieves better normalized distance. NICE identifies normoglycemic samples lying on the hyperglycemic side of the classifier for test sample a and b, while this is not the case for c.} 
\label{faq_explanation1}
\end{figure*}

\item Why does NICE achieve 100\% Cover, surpassing that of ExAct?

- NICE achieves 100\% Cover because the suggestions drawn by NICE are from the training set and they are always within the distribution of the training set.
    
\item Is it possible that the constrained intervention sometimes fails to deliver a counterfactual explanation that does not reflect user's preferences?

- Although rare, on few occasions the constrained intervention fails to comply with user preferences and eventually contributes to higher violations. 

\item For any dataset, are the external simulators same as the classifier?
    
- Definitely no, the external simulators are different models in terms of architecture. Even, the simulators are subject specific regression models for Nutrition Absorption and OhioT1DM. They forecast blood glucose levels given the counterfactuals. For Heart Disease and PimaDM, the simulators are subject agnostic models yet with different architectures.

\item Are CFs trustworthy for using as clinical interventions in diabetes management?

- The utility of CFs in clinical interventions is well-studied [2, 3]. The role of diet and medication in preventing hyperglycemia is well known. However, optimal change in these features for successful intervention is poorly understood. ExAct fills this gap by providing feedback on what changes are needed to prevent dysglycemia. Although we highlighted real-time usage of CFs following the outcome of the prediction model, it is important to note that our approach to generate CFs is independent of the prediction outcome. CFs can serve as educational interventions to the users through retrospective recommendations. The Functional Theory of Counterfactual Thinking [4] emphasizes the importance of thoughts about alternatives to past events and their role in behavior regulation and enhancing performance. ExAct can be used to improve glucose control by providing feedback on modifying past behaviors that could prevent hyperglycemia. Our current approach provides users with upward CFs (abnormality already took place, tell them how they could have prevented it). Similarly, ExAct can be used to provide downward CFs (normality took place, yet, tell them how it could have been abnormal).

\begin{figure*}[!h]
\centering
\includegraphics[width=0.7\linewidth,trim={1 6 1 2},clip]{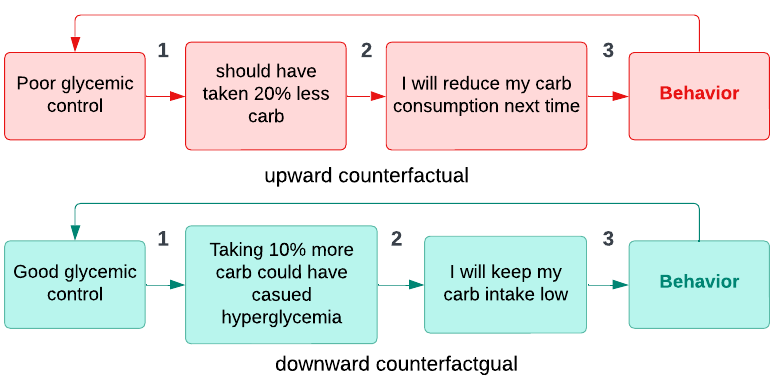}
\caption{Upward and downward counterfactual explanations.} 
\label{faq_explanation2}
\end{figure*}

\item Can multiple features have the same rank in the preference list?
    
- Yes, ExAct can accommodate features with the same rank. With little change to the grid search, the features can have the same rank in preferences. This way, features of the same rank are modified concurrently in an alternating process. Note that adding rank constraints may increase violations, depending on rank settings.

\item Are the prediction models biased? Which part of the pipeline is keeping the counterfactuals within the data manifold? 

- To keep the entire process free of bias, we randomly split the data to train and test. The bisection approach keeps the critical samples within the data manifold. The results in the paper (Plausibility, Table 1) show that not many CFs are outside of the data distribution.

\item If multiple CFs are generated, which one is to be used as  suggestion? Any pareto front has been used to identify the suggestion? Which constraint is given priority in this case?

- The CFs presented in the paper are the ones that minimally flips the class and are minimally distant from the test sample while preserving the preferences. Each point on the plot (Figure~\ref{trdof}) corresponds to a CF while all the CFs are generated for a single test (factual) sample. \textcolor{cyan}{$\times$} is the one we suggest in the paper i.e. we suggest the one minimally distant from the factual sample.

\begin{figure*}[!h]
\centering
\includegraphics[width=0.6\linewidth]{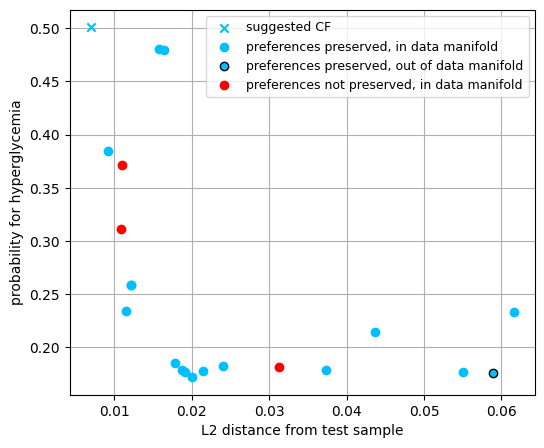}
\caption{Tradeoff plot for NA dataset. Each point corresponds to a CF. All the CFs are generated for a single test point.}
\label{trdof}
\end{figure*}

\item How this paper is different from state-of-the art (e.g. DiCE)?

Past studies also preserved user preferences within the generated counterfactuals (CFs); however, they treat user preferences as \textit{hard} constraints meaning that each input must be either included in or excluded from CFs. This can result in unnecessarily modifying certain inputs and creating interventions that although prevent an abnormal outcome, lead to another abnormal outcome due to drastic changes in input features. For example, when using DICE, we observed that the generated CFs suggest lowering pre-meal blood sugar level below 3 mmol/L resulting in dangerous hypoglycemic events. In contrast, we model user preferences as \textit{soft} constraints using a preference ranking approach. This rank-based accommodation of user preferences is entirely novel and has not been explored in the past. This way, we avoid making drastic changes because we first approximate decision boundary carefully using autoencoders and bisection method, ensuring that all CFs lie close to the critical region, then use preference ranks to ensure minimal change from the user's current behavior and avoiding potential risk. ExAct allows users to prioritize behavioral factors to remain unchanged by ranking them, offering more options and flexibility.

\end{itemize}

[1] Kevin Alex Zhang, Neha Patki, and Kalyan Veeramachaneni. Sequential Models in the Synthetic416 Data Vault. ArXiv, abs/2207.14406, 2022.

[2] Jonathan G. Richens, Ciarán M. Lee, and Saurabh Johri. Improving the accuracy of medical diagnosis with causal machine learning. Nature Communications, 11, 2020.

[3] Sandra Wachter, Brent Daniel Mittelstadt, and Chris Russell. Counterfactual Explanations Without Opening the Black Box: Automated Decisions and the GDPR. Cybersecurity, 2017.

[4] Kai Epstude and Neal J. Roese. The Functional Theory of Counterfactual Thinking. Personality and Social Psychology Review, 12:168 – 192, 2008.

\end{document}